\pdfoutput=1
\documentclass[review]{elsarticle}
\usepackage{subfigure}
\usepackage{longtable}
\usepackage{algorithm, algorithmic}
\usepackage{bm}
\usepackage{booktabs}
\usepackage{multirow}
\usepackage{geometry}
\usepackage{amsfonts,amssymb}
\usepackage{amsmath}

\usepackage{lineno,hyperref}
\modulolinenumbers[5]

\journal{arXiv}

\begin{document}

\begin{frontmatter}

\title{$\beta$-CapsNet: Learning Disentangled Representation for CapsNet by Information Bottleneck}


\author[mymainaddress]{Ming-fei Hu}
%\ead{luning_zhang@foxmail.com}

\author[mymainaddress]{Jian-wei Liu\corref{mycorrespondingauthor}}
\cortext[mycorrespondingauthor]{Corresponding author}
%\ead{liujw@cup.edu.cn}

\address[mymainaddress]{Department of Automation, College of Information Science and Engineering,
China University of Petroleum , Beijing, Beijing, China}

\begin{abstract}%摘要
 We present a framework for learning disentangled representation of CapsNet by information bottleneck constraint that distills information into a compact form and motivates to learn an interpretable factorized capsule. In our $\beta$-CapsNet framework, hyperparameter $\beta$ is utilized to trade-off disentanglement and other tasks, variational inference is utilized to convert the information bottleneck term into a KL divergence that is approximated as a constraint on the mean of the capsule. For supervised learning, class independent mask vector is used for understanding the types of variations synthetically irrespective of the image class, we carry out extensive quantitative and qualitative experiments by tuning the parameter $\beta$ to figure out the relationship between disentanglement, reconstruction and classfication performance. Furthermore, the unsupervised $\beta$-CapsNet and the corresponding dynamic routing algorithm is proposed for learning disentangled capsule in an unsupervised manner, extensive empirical evaluations suggest that our $\beta$-CapsNet achieves state-of-the-art disentanglement performance compared to CapsNet and various baselines on several complex datasets both in supervision and unsupervised scenes.
\end{abstract}


\end{frontmatter}

\section{Introduction}%正文

The disentangled representation can be specified as ones where single latent units are sensitive to changes in single generative factors, while being relatively invariant to changes in other factors \cite{1bengio2013representation}. If we could identity and separate out these factors, such representations distill information into a compact form which is often semantically meaningful and useful for standard downstream tasks such as supervised learning, transfer learning and reinforcement learning \cite{2ridgeway2016survey,3lake2016building}. There have been multiple efforts in deep learning towards learning disentangled representations, $\beta$-VAE \cite{4bengio2013representation} and InfoGAN \cite{5chen2016infogan} are significant methods for disentangling based on Variational Autoencoder (VAE) \cite{6kingma2014auto} and Generative Adversarial Networks (GAN) \cite{7goodfellow2014generative} framework. In addition to generative network, however, learning disentangled representations is difficult for some models due to the lack of effective constraints.

In this paper, we propose a framework based on Capsule Network (CapsNet) \cite{8sabour2017dynamic} and information bottleneck \cite{9tishby199937th,10tishby2015deep} that learns disentangled representation both in supervised manner and unsupervised manner. As a promising concept, CapsNet provides comparable performance on several benchmark datasets, it can be regarded as a special autoencoder whose representation is composed of some groups of neurons named capsule. However, it only can learn the entangled capsules that are unfavorable for most learning tasks. Therefore, in this paper, we intend to leverage information bottleneck to constrain the capsules for compressing representation space and learning disentangled factors.

Information bottleneck constraint is the intractable mutual information between the input and the representation, we present a variational bound to approximate the mutual information from the perspective of information theory. The variational bound of the mutual information is similar to the constraint of $\beta$-VAE, so our method is called $\beta$-CapsNet. In general, we can assume that the capsule vector in the representation is an isotropic unit Gaussian variable, the variance of capsule in the representation is related to the dimension, model structure and data type, therefore, we compel the mean of the capsule to 0.

In summary, we make the following contributions:

1) We introduce $\beta$-CapsNet, a novel approach for learning disentangled capsules constrained by information bottleneck, variational inference is used to construct an upper bound of information bottleneck constraint from the perspective of information theory, and the variational bound is tractable for most networks.

2) We proposed class independent mask vector to replace the existing mask matrix for understanding the types of variations synthetically irrespective of the image class for in a supervised manner, a series of quantitative and qualitative experiments show that our approaches can learn more interpretable representation, and grasp the relationship between disentanglement and other tasks by the trade-off parameter $\beta$.

3) We proposed the unsupervised $\beta$-CapsNet and the corresponding dynamic routing algorithm. Empirical evaluations suggest that our unsupervised $\beta$-CapsNet achieves state-of-the-art disentanglement performance compared to unsupervised CapsNet and various baselines on several complex datasets.

\section{Related Work}

\textbf{Disentangled representations:}Early works to attempt to learn disentangled latent factors include punishing predictability of certain latent dimension in auto-encoder \cite{12schmidhuber1992learning} and Boltzmann machine \cite{13desjardins2012disentangling}. More recent works have focused on modeling the variation factors of generative model such as InfoGAN and $\beta$-VAE. InfoGAN maximizes the mutual information between a small subset of the latent variables and the observation \cite{5chen2016infogan}, our $\beta$-CapsNet minimizes the mutual information between the input and the representation. $\beta$-VAE \cite{4bengio2013representation} uses a modified version of the VAE objective with a larger weight ($\beta$ \textgreater1) on the KL divergence, our $\beta$-CapsNet adopt a similar objective with the hyperparameter $\beta$ to compress the representation space. $\beta$-TCVAE carries out a decomposition of the variational lower bound \cite{14chen2018isolating} and uses the total correlation or mutual information term \cite{15achille2018information} to explain the success of $\beta$-VAE in learning disentanglement. FactorVAE encourages the code distribution to be factorial by using a discriminator that distinguishes whether the input was drawn from the marginal code distribution or the product of its marginals \cite{16kim2018disentangling}, it’s an ingenious combination of $\beta$-VAE and GAN for disentanglement. Joint-VAE learns disentangled jointly continuous and discrete representations for disentangling the factors of different categories on supervised data \cite{17dupont2018learning}. Different approaches have been explored for semi-supervised or supervised learning of disentangled representations \cite{18kulkarni2015deep,19yang2015weakly,20reed2014learning}, however, most previous attempts are based on generative model, and there is no effective method to learn disentangled representations for other models such as CapsNet.

\textbf{Information bottleneck:}The definition of information bottleneck is proposed in \cite{9tishby199937th}, using this objective for deep neural networks is pointed out in \cite{10tishby2015deep,21shwartz2017opening} but no including the verification experimental results. Deep variational information bottleneck constructs the lower bound of the information bottleneck objective in high-dimensional continuous neural network through variational inference \cite{11alemi2017deep}, it has been successfully applied in deep learning for better representations \cite{22peng2019variational,23chalk2016relevant,24federici2020learning}. Information dropout injects multiplicative noise in the activations of neural network to approximate information bottleneck constraints \cite{25achille2018information}, and a similar algorithm is used to limit the primary capsules of CapsNet for better performance and less computation \cite{26hu2021learning}. However, to our best knowledge, learning disentangled representations through information bottleneck constraint is the first attempt.

\textbf{Capsule network:}The research history of invariant spatial relationships between the object and its parts can be dated back to \cite{27hinton1981parallel}, the notion of capsule \cite{28hinton2011transforming} and dynamic routing \cite{8sabour2017dynamic} package this theory into CapsNet, which is regarded as the first theoretical prototype. Each capsule represents an instance of an entity composing of several neurons, and dynamic routing is an iterative mechanism to send lower-level capsules to higher level. Most works pay attention to novel versions of capsules \cite{29hinton2018matrix}, faster dynamic routing algorithms \cite{30ribeiro2020capsule,31dou2019dynamic,32wang2018optimization,33li2018neural} and deeper layers \cite{34kosiorek2019stacked,35rajasegaran2019deepcaps}, however, the algorithms for discovering disentangled capsule are open problem.

\section{Learning Disentangled Capsules by Information Bottleneck}

In this section, we would embark on a discussion of learning disentangled capsules by information bottleneck constraint. Firstly, we recall the conceptions of capsule, dynamic routing and reconstruction network briefly, and we propose the idea of constraining the representation of CapsNet by information bottleneck that encourages the network to learn disentangled factors. Secondly, we introduce information bottleneck algorithm and discuss how to use it as an additional information loss to constrain the representation of CapsNet, therefore a novel framework $\beta$-CapsNet is proposed. We attend to address the intractable mutual information in the loss, then variational inference derivation is presented to construct a variational upper bound. We assume that the prior is centered isotropic multivariate Gaussian, and we parametrize the posterior by a factorized Gaussian which mean and variance depend on the CapsNet’s representation, then the information loss can be integrated analytically in a simple form. Lastly, class independent mask vector is proposed for understanding the types of variations synthetically irrespective of the image labels. Our mask vector, which only sends the correct capsule to the decoder, instead of all classified capsules, it can force decoder to learn jointly disentangled representation with the same parameters. In addition, our decoder network consisting of deconvolutional layers can capture more spatial relationships from the complex input images and reconstruct clear images.

\subsection{A Brief Conception about CapsNet}

Capsule consisting of a group of neurons is the essential unit in the CapsNet that represents the instantiation parameters of a specific type of entity such as an object or an object part \cite{8sabour2017dynamic}, higher-level capsules represent more complex entities with more degrees of freedom. Fig.1 illustrates a brief architecture of CapsNet, the primary capsules provide the lower-level of multi-dimensional entities, the classified capsules which are long instantiation vectors for inputs are the highest-level representations for an object or an object part, these lengths are used to represent the probability and the one with the longest length is predicted result for classification task. 

The lower-level capsules are sent to higher-level classified capsules by a very different type of computation named dynamic routing. As an iterative routing-by-agreement mechanism on supervised datasets, routing algorithm assigns capsules depending on whose activity vectors have a big scalar product. Routing mechanism ensures that the classified capsules can predict input’s class label by length, and the corresponding representation can obtain enough information from features for reconstructions, however, a novel routing is needed for unsupervised learning.

\begin{figure}[!htbp]
	\centering
	\includegraphics[scale=1]{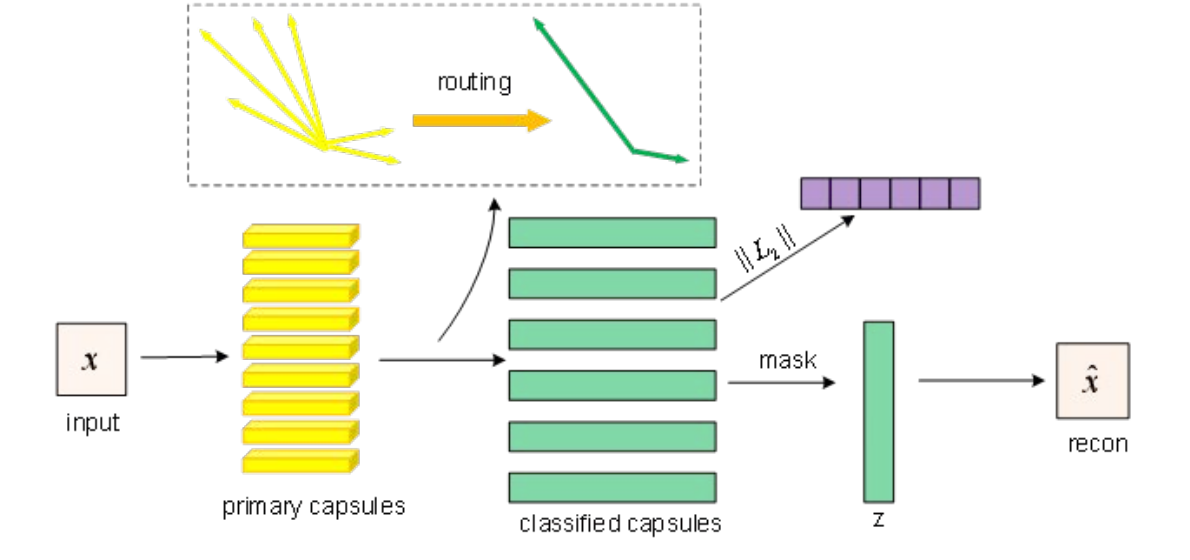}
	\caption{The brief architecture of CapsNet}
	\label{fig1}
\end{figure}

The reconstruction network utilizes the regularization method to alleviate the overfitting and boost the accuracies on some basic classification datasets. During training, mask matrix is used to mask out all with zeros but the capsule of the correct label, and these new classified capsules are flattened to a vector as the representations of CapsNet. Reconstruction network can reconstruct images from the representations while keeping important details. 

Unfortunately, the numerous factors in representations generated by above approaches are highly complex interaction with others. We guess the reason CapsNet’s entangled representation could be that no suitable constraints are incorporated to restrict the generating process for the representation. In order to retain all the valuable information, CapsNet has no feature compression process such as pooling layer, it causes that the learned representation don’t identify the salient features underlying in objects with significant differentiation. Therefore, imposing appropriate constraints on the generating process for representation without damaging any valuable information underlying in objects is the key point to learning disentangled representations of CapsNet. 

To discover the independent latent factors of variation, and explore the relationship between disentangled representations, reconstructions and classification accuracies of CapsNet, we introduce the information bottleneck constraint, an additional term in loss function to abstract the relevant information of representation and encourage the network to discover disentangled factors. 

\subsection{The Loss Function of $\beta$-CapsNet}

In this subsection, we would focus on formalizing the ideas of learning disentangled capsules by information bottleneck. Given some input data x and the representation denotes as z, information bottleneck suggests that constraining the mutual information between x and z can able to compress representation space and enhance the interpretability of the representations in perspective of information theoretic concepts, it is equivalent to solve the optimization problem:

\begin{equation}
\begin{array}{l}
\min \;\;I(x;z)\\
{\rm{s}}{\rm{.t}}{\rm{.}}\;\;\;I(x;y) = I(z;y)
\end{array}
\end{equation}

where  $I( \cdot )$ denotes the mutual information. The corresponding Lagrangian dual formulation for Eq. (1) is written as follow in literature \cite{9tishby199937th}:

\begin{equation}
{L_{{\rm{IB}}}} =  - I(z;y) + \beta I(x;z)
\end{equation}

where $\beta$ is positive constant. Most researches use the second term as a regularization term because the first term can be replaced by the loss function from original problem. Then we can get the loss function, that corresponds to classification loss, reconstruction loss and disentanglement constraint respectively, as follow:

\begin{equation}
L = {\sum L _k} + \alpha {L_{{\rm{mse}}}} + \beta I(x;z)
\end{equation}

Mutual information is a fundamental quantity leverage for measuring the relevant information between two variables:

\begin{displaymath}
I(x;z) = \int {\int {p(x,z)\log \frac{{p(x,z)}}{{p(x)p(z)}}dzdx} } 
\end{displaymath}

where $p(x,z)$ is the joint probability distribution, $p(x)$ and $p(z)$ are the marginals, any one of them is difficult to compute for our model, it is necessary to devise some simple methods to estimate it. Here we use variational inference to estimate the mutual information while constructing a variational bound formulation. In information theory, mutual information can be seen as the uncertainty in x given z:

\begin{displaymath}
%I(x;z) ={ H(z) - H(z|x)\\
% - \int {p(z)\log p(z)dz} + \iint {{p(x,z)\log p(z|x)}dxdz}}
\begin{array}{l}
I(x;z) = H(z) - H(z|x)\\
\;\;\;\;\;\;\;\;\; =  - \int {p(z)\log p(z)dz}  + \iint {{p(x,z)\log p(z|x)}dxdz}
\end{array}
\end{displaymath}

where $H( \cdot )$ denotes the Shannon entropy and $H(x|z)$ denotes the conditional entropy. Directly computing the marginal distribution $p(z) = \int {p(z|x)p(x)dx} $ is difficult, so let $q(z)$  be a variational approximation to this marginal:

\begin{displaymath}
\begin{array}{l}
\;\;\;\;\int {p(z)\log p(z)dz}  \approx {\rm{KL}}\left( {p(z){\rm{||}}q(z)} \right) \ge 0\\
 \Rightarrow \int {p(z)\log p(z)dz}  \ge \int {p(z)\log q(z)dz} 
\end{array}
\end{displaymath}

Then we have the following variational upper bound:

\begin{displaymath}
\begin{array}{l}
I(x;z) = \iint{{p(x,z)\log \frac{{p(z|x)}}{{p(z)}}}dxdz}\\
\;\;\;\;\;\;\;\;\;\; \le   \iint {{p(x,z)\log \frac{{p(z|x)}}{{q(z)}}}dxdz}\\
\;\;\;\;\;\;\;\;\;\;  \approx \int {p(x)} \left( {\int {p(z|x)\log \frac{{p(z|x)}}{{q(z)}}dz} } \right)dx\\
\;\;\;\;\;\;\;\;\;\; \approx \frac{1}{N}\sum\limits_{n = 1}^N {{\rm{KL}}\left( {p(z|{x_n})||q(z)} \right)} 
\end{array}
\end{displaymath}

Naturally, we can assume the prior is normal distribution
 $p(z) \sim N(0,I)$%式子修改
, where $I$ denotes identity matrix. The posterior has the form $p(z|x) = N(\mu ,{\sigma ^2})$ where $\mu $ and ${\sigma ^2}$ denote mean and variance respectively, these parameters can be constructed from encoder of CapsNet. Then combining all the facts derived, we have:

\begin{equation}
\begin{array}{l}
I(x;z) \approx \frac{1}{N}\sum\limits_{n = 1}^N {{\rm{KL}}\left( {p(z|{x_n})||q(z)} \right)} \\
\;\;\;\;\;\;\;\;\; = \int {N(\mu ,{\sigma ^2})} \log N(\mu ,{\sigma ^2})dz - \int {N(\mu ,{\sigma ^2})} \log N(0,{\rm{I}})dz\\
\;\;\;\;\;\;\;\;\; = \frac{1}{2}\left( {{\mu ^2} + {\sigma ^2} - \log ({\sigma ^2}) - 1} \right)
\end{array}
\end{equation}

In this case, the KL divergence term controls the degree of disentanglement by twiddling the coefficient $\beta$, it is similar to the $\beta$-VAE’s loss function that has evaluated the regularization term on unsupervised learning, so we call our model $\beta$-CapsNet. $\beta$-CapsNet limits the relevant features of the representation and forces the representation to complete the reconstructions with fewer but more general features, therefore, our disentangled representation would remove the complex details and highly interaction latent factors.

CapsNet use the length of classified capsule to represent the probability of belonging to a certain class, hence the separate margin loss ${L_k}$ for classified capsule k in multi-classification scenario is formulated as follow:

\begin{displaymath}
{L_k} = {T_k} \cdot f{(0.9 - {l_k})^2} + 0.5(1 - {T_k}) \cdot f{({l_k} - 0.1)^2}
\end{displaymath}

where ${T_k} = 1$ when class k is present, $f$ denotes rectified linear unit and ${l_k}$ is the length of classified capsule k. 

The reconstruction network reconstructs the input image from the classified capsule to encourage the representation to encode the instantiation parameters, its goal is that the reconstruction is as similar to the input image as possible. So the additional reconstruction loss ${L_{{\rm{mse}}}}$ is Euclidean distance between input and reconstruction:

\begin{displaymath}
{L_{{\rm{mse}}}} = |x - \hat x{|^2}
\end{displaymath}

\subsection{Class Independent Mask Vector for Supervised Data}

The decoder in CapsNet, which is consisted of three fully connected layers, is class dependent. For supervised learning, we assume that the mask matrix $M \in \mathbb{R}^{a \times b}$ denotes the activity matrix of mask for all classes, where a is the number of classes and b is the capsule dimension. As illustrated by Fig.2, classified capsules are masked by activity vector M with label y and other capsules are masked by zeros, it results in M as shown follow:

\begin{displaymath}
{m_i} = \left\{ {\begin{array}{*{20}{c}}
{{p_i}}\\
0
\end{array}\;\;\;\begin{array}{*{20}{c}}
{i = y}\\
{i \ne y}
\end{array}} \right.
\end{displaymath}

where $i$ is $i$-th class of capsules. After masked by matrix M that provides class information to the decoder indirectly, the decoder becomes class dependent, then the capsules are flattened as a one-dimensional vector and fed into the decoder network.

\begin{figure}[!htbp]
	\centering
	\includegraphics[scale=1]{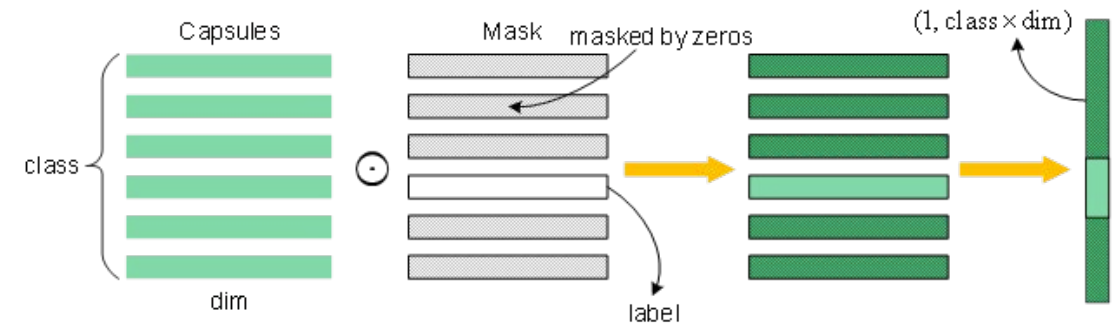}
	\caption{Class dependent mask matrix}
	\label{fig2}
\end{figure}

However, a significant limitation of class dependent mask is that the latent factors captured by instantiation parameters lack controllability and interpretability. For example, one disentangled factor for a given class causes some style variable, there is no guarantee that the same dimension would cause same style in other classes, because capsules in different classes maybe have different positions and parameters. As a result, it is really a challenge for learning disentangled latent representations.

Hence, we propose a class independent mask vector for disentangled capsules and better reconstruction images. Instead of building mask matrix for all-classes, we only send the correct one to the decoder as shown by Fig.3. Let  ${\rm{M'}} \in \mathbb{R}^{1 \times b}$ denote the activity vector of mask for class label, the mask vector can force decoder to learn disentangled representation jointly within a constrained space and same parameters, the instantiation parameters and interpretable latent factors of all classes are learnt from the same distribution and the same dimension of input vector. 

\begin{figure}[!htbp]
	\centering
	\includegraphics[scale=1]{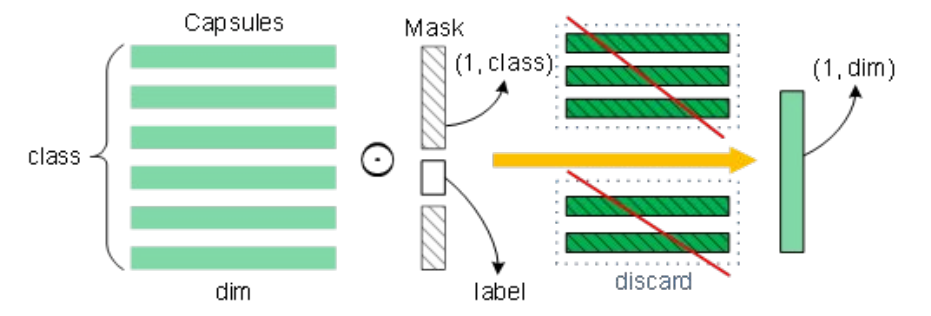}
	\caption{Class independent mask vector}
	\label{fig3}
\end{figure}

Our mask vector helps us to understand the types of variations synthetically irrespective of the image class. Furthermore, it can also learn more interpretable properties such as a variation from one class to another. In addition, in order to obtain clearer and sharper reconstructions, we replaced the fully connected layers of the decoder with some deconvolution layers.

\section{The Specific Implementation of $\beta$-CapsNet}

In this section, we would describe the specific implementation of $\beta$-CapsNet on supervised learning and unsupervised learning setup, respectively. Firstly, we introduce $\beta$-CapsNet in three parts: encoder, decoder and representation, then we demonstrate that the variance of the representations is related to the dimension and a new constraint algorithm that limits the mean to 0 is presented to replace the KL term. Secondly, we show how to use $\beta$-CapsNet to learn disentangled representation in a supervised manner, where class independent mask vector is used to capture more controllable and interpretable latent factors. Lastly, we introduce a novel dynamic routing algorithm for unsupervised learning, and then we describe the specific implementation of unsupervised $\beta$-CapsNet.

\subsection{The Structure of Encoder, Decoder and Representation}

If the CapsNet is viewed as an auto-encoder, the convolutional layers and capsule layers can be regarded as encoder, the reconstruction network is corresponding to decoder, the classified capsule vector after masking operation is the representation. In the following, we will discuss the encoder, the decoder and the representation shown in Fig.4, respectively.

\textbf{Encoder:} As shown in Fig.4 (a), some convolutional layers are used to convert pixel intensities to the activities of local features from input without pooling layer. These features named blocks are divided into m primary capsules and each primary capsule is an 8D vector. The final layer has one 16D classified capsule per class, each of them receives input from all primary capsules through dynamic routing. The size of weight matrix W in routing is $(m,8,16 \times n)$  ,where $n$ denotes class, this weight can achieve a better initialization of the routing and change the dimensionality from primary capsules to classified capsules.

\begin{figure}[!htbp]%图片
	\label{fig4}
	\centering
	\subfigure[Encoder structure]{
		\includegraphics[width = .7\textwidth]{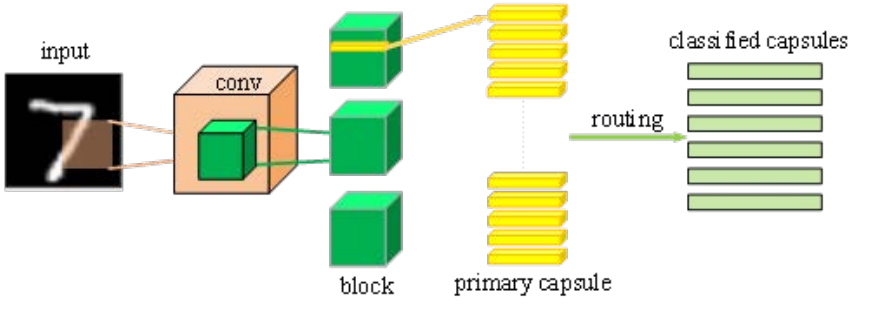}
	}
	\quad
	\subfigure[Decoder structure]{
		\includegraphics[width = .7\textwidth]{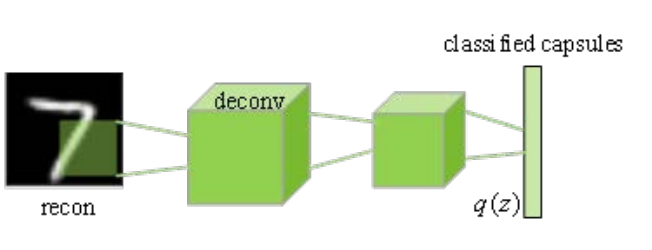}
	}
	\quad
	\subfigure[The construction of disentangled representation]{
		\includegraphics[width = .7\textwidth]{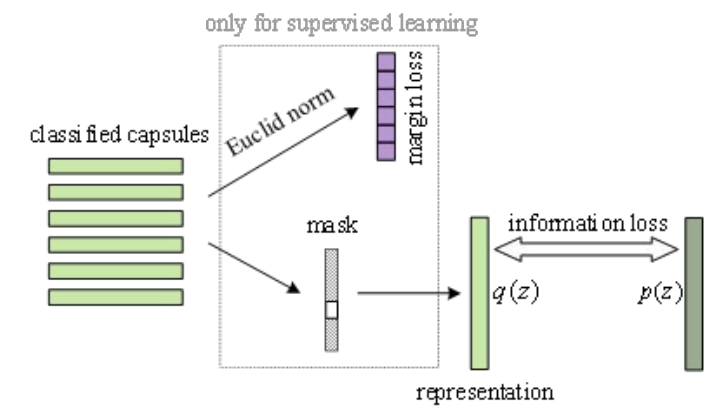}
		%\caption{fig1}
	}
	\caption{The construction of the encoder, the representation and the decoder}
\end{figure}

\textbf{Decoder:} As shown in Fig.4 (b), the existing decoder consists of three fully connected layers that is suited for simple datasets such as MNIST. We replace the decoder with a deconvolutional network which is better at reconstructing spatial relationships and instantiated entities when the input images are complex. We find that the batch normalization will affect disentanglement in CapsNet, it is different with VAE such as $\beta$-TCVAE \cite{14chen2018isolating}.

\textbf{Representation:} The construction process of disentangled representation is shown in Fig.4 (c). Classified capsules have two roles in supervised learning: the first one is classification, according to the Euclid norm of the classified capsules we can get the length vectors, and then we use the vectors to calculate margin loss. The second function is reconstruction, classified capsules masked by class independent mask vector is our representation, and then we send the representation to the decoder for reconstructing. In an unsupervised manner, we don’t need to consider impact of class label, and our mask vector should be removed because there is only one capsule vector in the representation.

In general, we can assume that the output of encoder is the variance of $p(z|x)$. In order to let the length of classified capsule represent the probability of the entity occurrence in the current input, dynamic routing mechanism contains a novel non-linear function named ‘squash’ for lower-level capsules in Eq. (5) and a ‘routing softmax’ for the coupling coefficients between capsules in Eq. (6):

\begin{equation}
f(v) = \frac{{||v|{|^2}}}{{1 + ||v|{|^2}}}\frac{v}{{||v||}}
\end{equation}

\begin{equation}
{c_{ij}} = \frac{{\exp ({b_{ij}})}}{{\sum\nolimits_j {({b_{ij}})} }}
\end{equation}

where ${b_{ij}}$ denotes coefficient between capsule $i$ and capsule $j$ in higher-level layer. These functions ensure that the longest classified capsule get shrunk to a length slightly below 1 and other lengths get shrunk to almost zero. However, constraining the variance is an unreasonable assumption for our representation due to squash activation and softmax function, the variance is related to the dimension and $\beta$ as shown in Fig.5. Therefore, it is necessary to build an alternative constraint that is more suitable for CapsNet.

\begin{figure}[!htbp]
	\centering
	\includegraphics[scale=0.5]{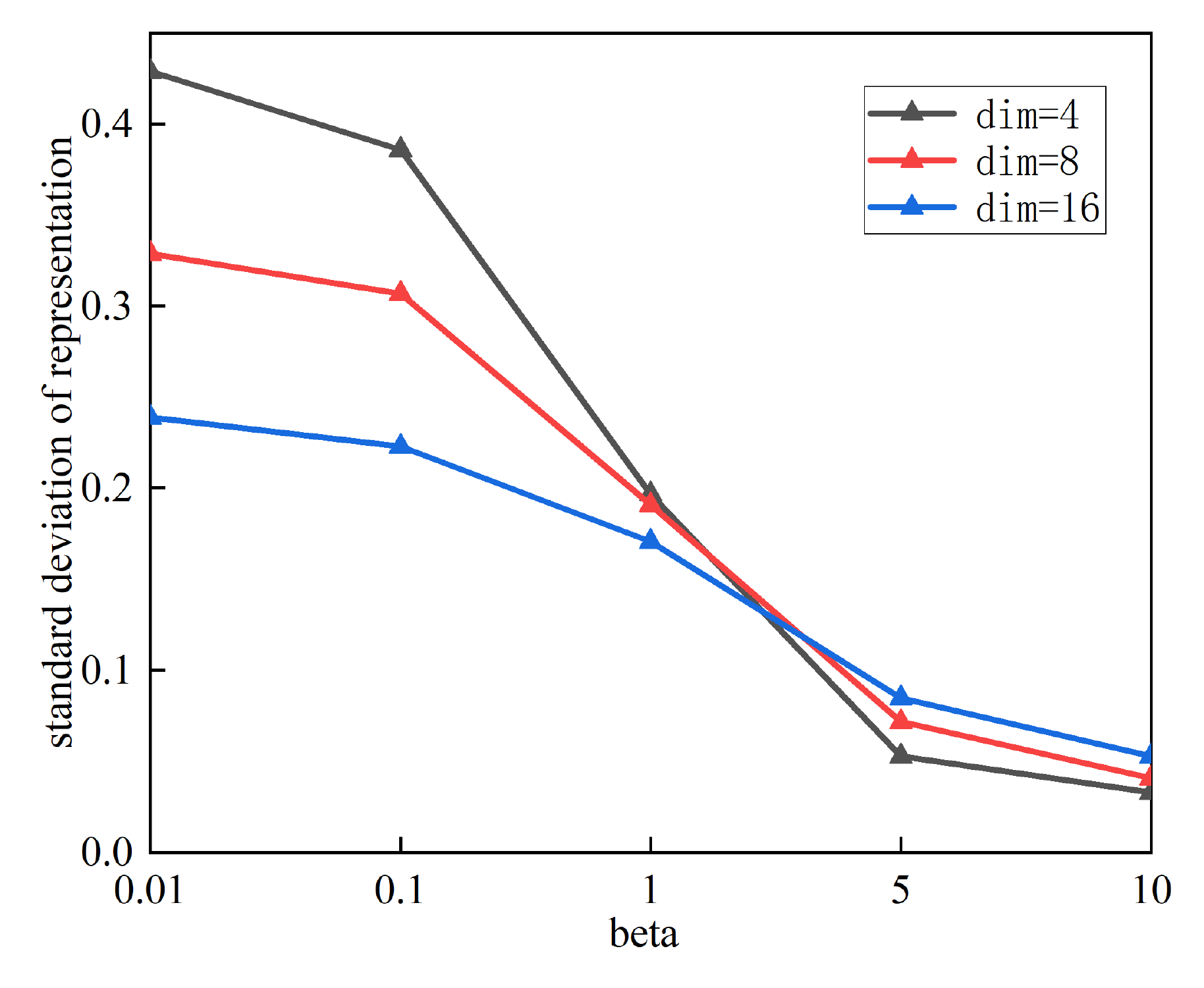}
	\caption{ For three different dimensions (4, 8, 16) of the representation, we show the plot of the mean of standard deviation trained on MNIST with five different values of $\beta$ (0.01, 0.1, 1, 5, 10). A large value of $\beta$ indicates that the information bottleneck term has a strong constraint on the representation. While for low value of $\beta$, the standard deviation of the representation is larger. In addition, the standard deviation is related to the dimensions of capsule, when $\beta$ is less than 1, more neurons in representation will decrease its variance, and the relationship tends to reversed when $\beta$ is larger than 1.}
	\label{fig5}
\end{figure}

To tackle above problem, a new constraint is proposed: we assume that the output of the encoder is the mean of $p(z|x)$ and set the mean to approximate equal to 0, the constraint on the variance is abandoned due to its indeterminacy. It is a simpler and more appropriate choice of space compression that only the mean of the capsules needs to be limited. 

\begin{displaymath}
\begin{array}{l}
\;\;\;{\rm{KL}}\left( {p(z|x)||q(z)} \right)\\
 = \frac{1}{2}\left( {{\mu ^2} + {\sigma ^2} - \log ({\sigma ^2}) - 1} \right)\\
 = \frac{1}{2}\left( {{\mu ^2} - 1} \right)
\end{array}
\end{displaymath}

Now we can get final loss function as:

\begin{equation}
L = {\sum L _k} + \alpha {L_{{\rm{mse}}}} + \frac{\beta }{2}\left( {{v^2} - 1} \right)
\end{equation}

where $v$ denotes the representation after masked matrix.

\subsection{$\beta$-CapsNet in a Supervised Manner}

In supervised learning, the datasets we used are MNIST and Fashion-MNIST, so we assume that the input size of $\beta$-CapsNet is $({\rm{1,28}},{\rm{28}})$, its structure is shown in Fig.6. In this case, we would follow the encoder settings of CapsNet, except that the number of filters in second convolutional layer has been adjusted. The first convolutional layer has 256 filters, 9 kernels, 1 stride and ReLU activation, the second layer has 128 filters, 9 kernels and a stride of 2. A capsule block (the size is $(8,6,6)$ ) contains the output of 8 filters, each block has 36 primary capsule vectors and each of them is an 8D vector, all the vectors in same block are sharing their weights with each other.

\begin{figure}[!htbp]
	\centering
	\includegraphics[scale=1]{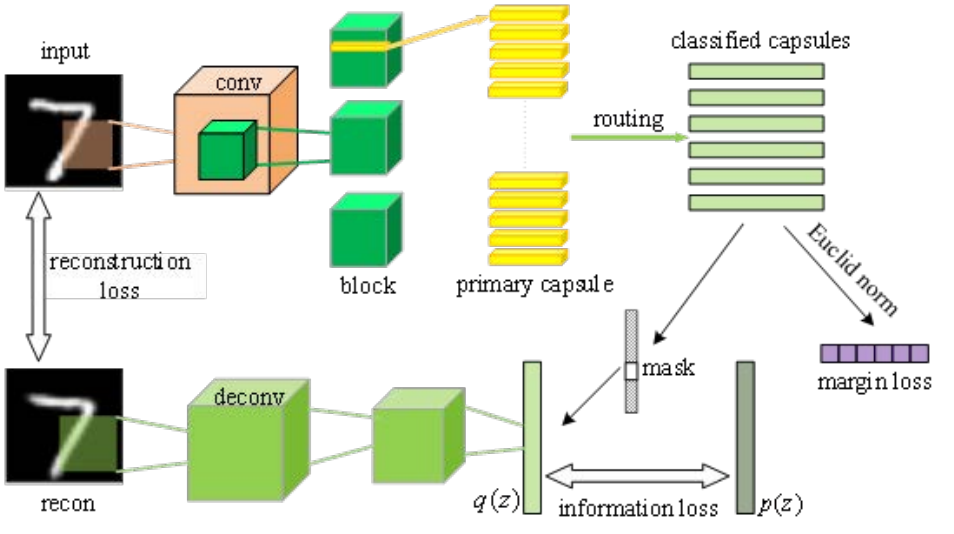}
	\caption{The structure of $\beta$-CapsNet constrained by information bottleneck}
	\label{fig6}
\end{figure}

In total, primary capsules layer has ${\rm{16}} \times {\rm{6}} \times {\rm{6}}$ capsule outputs. The next layer named classified capsule has one 8D capsule per class and each of these capsules receives input from all the primary capsules, the size of the weight matrix in dynamic routing is (576, 8, 80). After masking by our mask vector, our representation (an 8D vector) is limited by information bottleneck constraint, the information loss between representation distribution $q(z)$ and standard normal distribution $p(z)$ in Eq. (7) is an additional loss item in whole loss function.

Five deconvolutional layers are used in our decoder, the detailed structure of $\beta$-CapsNet and CapsNet are exhibited in Table 1. There are two kinds of decoders in our model: fully connected network is used for simple dataset such as MNIST and deconvolutional network is suitable for complex images such as Fashion-MNIST.

\begin{table}[!htbp]%表 multirow合并行  multicolumn合并列
	\centering
	\label{tb1}
	\caption{Structure Comparison of $\beta$-CapsNet and CapsNet in supervised learning}
	\begin{tabular}{c|cc}
		\hline
	%	\multirow{2}{*}{Dataset} & \multirow{2}{*}{C} & \multicolumn{3}{c}{Sequence   Length} & \multirow{2}{*}{Events} \\ \cline{3-5}
         model	&$\beta$-CapsNet	&CapsNet \\ \hline
  input	& \multicolumn{2}{c}{(1, 28, 28)} \\
\multirow{2}{*}{encoder}  &\multicolumn{2}{c}{Conv (256, 9×9, 1)} \\
   &Conv (128, 9×9, 2) & Conv (256, 9×9, 2)\\
primary capsule   &(576, 8)	&(1152, 8) \\
routing matrix  & (576, 8, 80)	&(1152, 8, 160) \\
classified capsule	&(10, 8)	&(10, 16) \\
mask	&mask vector	&mask matrix \\
representation	&(1,8)	&(1,160) \\
decoder	&Deconv (256, 4×4, 1)	&FC (512) \\
	&Deconv (128, 4×4, 2)	 &FC (1024) \\
	&Deconv (64, 9×9, 1)	&FC (784) \\
	&Deconv (32,9×9, 1)	&\\
	&Deconv (1, 9×9, 1)	& \\
output	 &\multicolumn{2}{c}{(1, 28, 28)}  \\ \hline
	\end{tabular}
\end{table}

There are some issues worth discussing in our setting. First, there is fewer capsule blocks and primary capsules, it can greatly reduce the computational complexity without affecting the reconstruction and learning disentangled representation. Second, the classified capsules are 8D vectors instead of 16D because 8 dimensional is the most suitable setting for reconstruction and disentanglement as shown in Fig.7. Therefore, our representation is an 8D vector and the dimension of the representation in CapsNet \cite{8sabour2017dynamic} is 160 (most of them are masked by zeros). 

\begin{figure}[htbp]%多图片
\centering
\subfigure[Train images of MNIST with different dimensions of representation]
{
    \begin{minipage}[b]{.8\linewidth}
        \centering
        \includegraphics[scale=0.6]{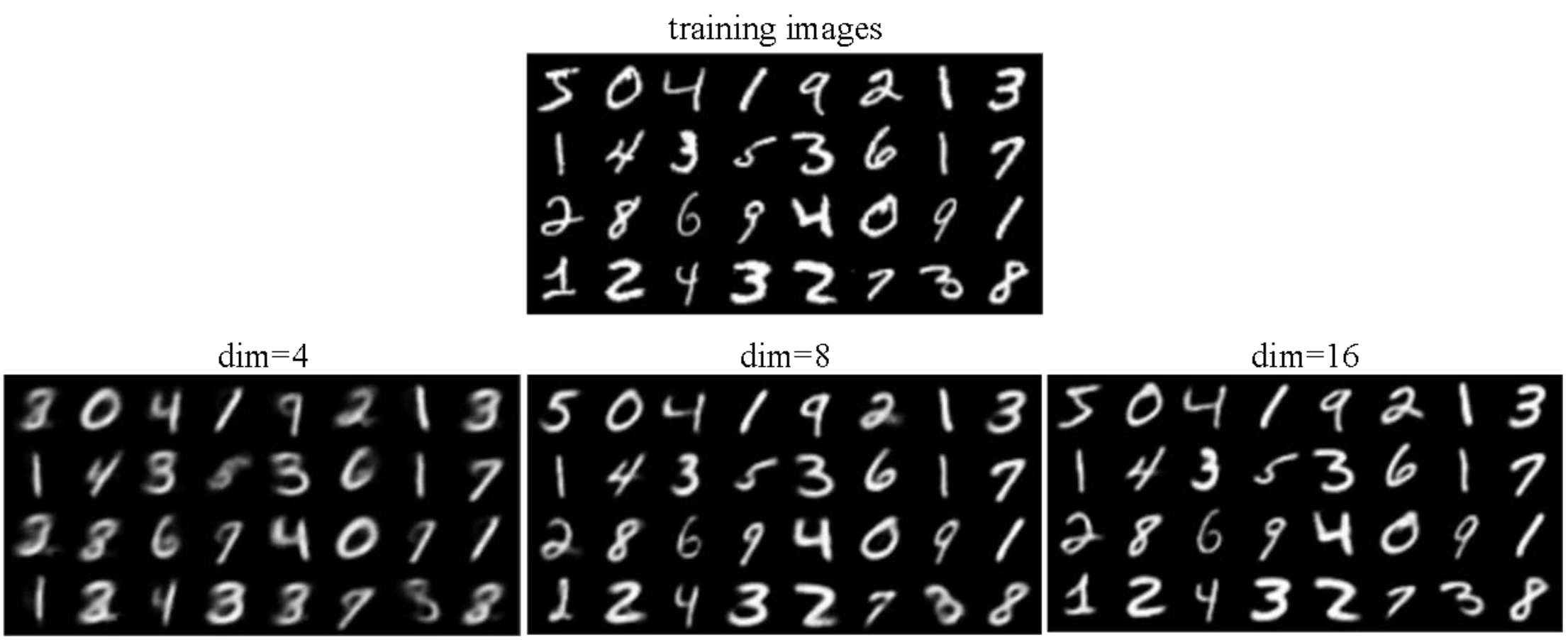}
    \end{minipage}
}
\subfigure[Test images of MNIST with different dimensions of representation]
{
 	\begin{minipage}[b]{.8\linewidth}
        \centering
        \includegraphics[scale=0.6]{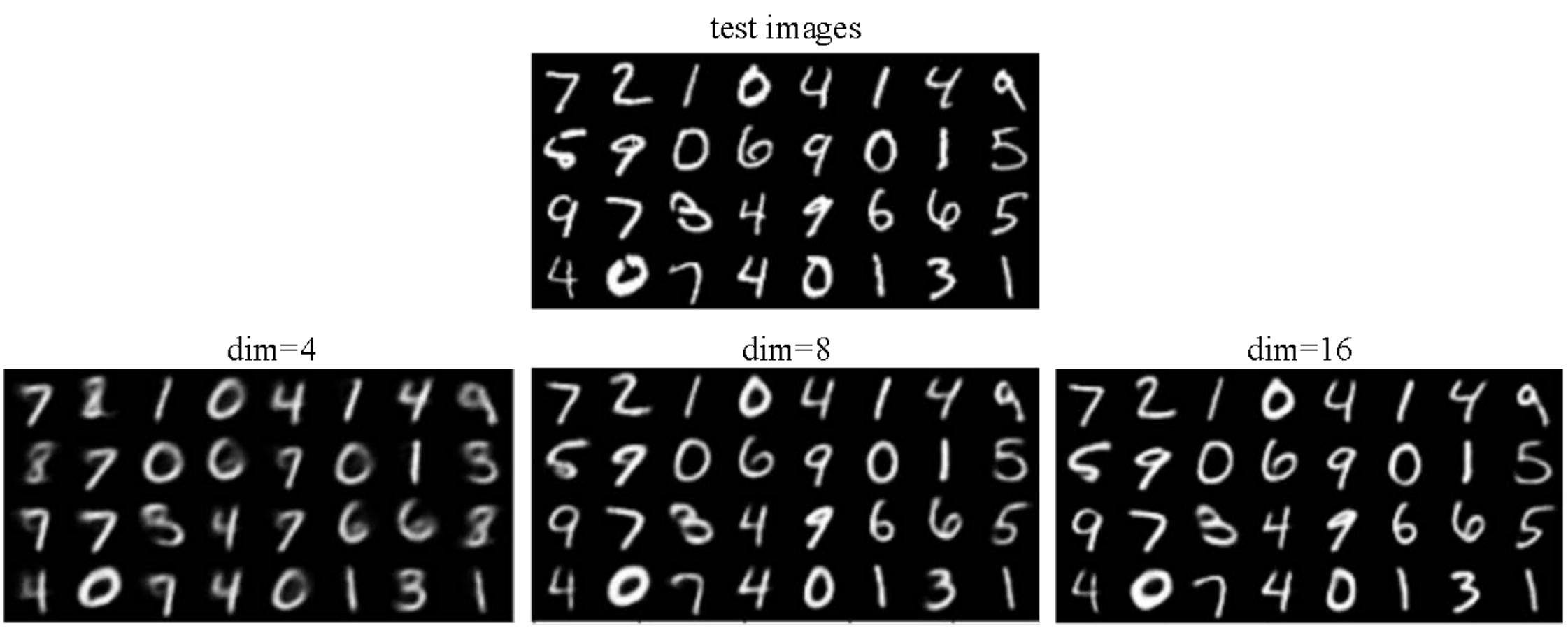}
    \end{minipage}
}
\caption{Reconstructed images on MNIST \cite{36lecun1998mnist} with fully connected decoder ($\beta$=3). (a) In training dataset, the reconstructions with dim=16 have high fidelity, most of the details are preserved, indicating that the capsule contains enough neurons to recover the input images. In contrast, the reconstructions with dim=4 are serious distortion. (b) In test dataset, the fidelity of reconstructions with dim=8 is very close to dim=16. Compared with the training dataset, the decoder cannot retain enough details in the reconstruction during test, it is an ‘overfitting’ in reconstruction task.}
\end{figure}

\subsection{$\beta$-CapsNet for Unsupervised Data}

CapsNet is used for the classification task in supervised learning, we need to modify the model structure and routing algorithm so that the model can handle unsupervised data, then we will describe the specific implementation of unsupervised $\beta$-CapsNet.

\textbf{Unsupervised Structure:}Compared with CapsNet in supervised learning scene, the unsupervised model has two characteristics. Firstly, unsupervised samples have no labels, the related calculation process can be deleted such as softmax function and margin loss function. Secondly, we can regard unsupervised samples as supervised data with only one class, the number of classified capsules is 1 that is the representation of unsupervised CapsNet. Therefore, the mask vector we proposed should be removed because there are no more redundant capsule vectors in the representation and the representation of all samples is the same capsule vector.

\textbf{Unsupervised Dynamic Routing:} In the supervised manner, dynamic routing in \cite{8sabour2017dynamic} assigns the instantiation features of capsule vectors to all capsules of next layer, we can use it to pass the features and information in capsule layers. Therefore, a slight modification is needed for unsupervised data to merge all capsules into the last capsule vector because the last capsule layer contains only one capsule. 

We iterate through the proposed routing algorithm r times which is set to 3 empirically following \cite{8sabour2017dynamic}. The inputs of unsupervised routing are capsule vectors u in layer $l - 1$ and weight matrix ${\rm{W}}$, ui and ${{\rm{W}}_i}$ denote the $i$-th capsule vector and corresponding weight; after the iterations, the output is capsule vector v in last capsule layer 
$l$. In the insitialized procedure, ${\hat u_i}$ is a prediction vector after spatial mapping and dimensional transformation which is produced by multiplying the $i$-th capsule vector ui by the weight ${{\rm{W}}_i}$, ${b_i}$ and ${c_i}$ are the log prior probabilities and coupling coefficients between capsule ${u_i}$ and $v$.

\begin{algorithm}
	\caption{Unsupervised dynamic routing algorithm}
	\label{alg1}
	\begin{algorithmic}[1]
%		\REQUIRE The number of encoding layers: $n$, event-type encoding $({\bm{EC}}_n )^T$
%		. Temporal encoding (position encoding) ${\bm{X}}^T$ .
%		\ENSURE Hidden representation (State) $	{\mathbf{H}} \in \mathbb{R}^{D \times I_N } $of event sequence  
		\STATE  \textbf{procedure} ROUTING
		\STATE \textbf{Require:} $r$, ${u_i}$, ${{\rm{W}}_i}$, $l$
		\STATE  \textbf{Initialize:} \\${\hat u_i} = {{\rm{W}}_i}{u_i}$\\${b_i} \leftarrow 0$\\${c_i} \leftarrow {\rm{soft}}\max ({b_i})$\\$v \leftarrow {\rm{squash}}\left( {\sum\nolimits_i {{c_i}{{\hat u}_i}} } \right)$
		\STATE \textbf{for} r iterations \textbf{do}\\
               for all capsule i in layer $l - 1$ and capsule  v in layer l: ${b_i} \leftarrow {b_i} + {\hat u_i} \cdot v$\\ 
  for all capsule i in layer $l - 1$: ${c_i} \leftarrow {\rm{soft}}\max ({b_i})$\\
$v \leftarrow {\rm{squash}}\left( {\sum\nolimits_i {{c_i}{{\hat u}_i}} } \right)$\\
\textbf{return} v
	\end{algorithmic}
\end{algorithm}

Fig.8 shows the structure of unsupervised $\beta$-CapsNet that contains two capsule layers and an unsupervised routing algorithm. The largest unsupervised data we used is CelebA, so we set the input size as $({\rm{3,64}},{\rm{64}})$  (after resizing) to analyze the specific structure. In this case, four convolutional layers are used to construct primary capsules, more hidden layers can help to extract more advanced features and reduce the number of primary capsules. There are 576 primary capsule vectors (each vector is an 8D vector) in the first capsule layer and a weight matrix (576, 8, 16) in unsupervised routing. The output of routing is the representation which is a 16D vector, six deconvolution layers are used to reconstruct samples that can properly capture the spatial relationships from the representations, the detailed structure of encoder and decoder used in the experiments are exhibited in Table 2. We refer to the hyperparameters and settings of convolutional layers and deconvolutional layers of $\beta$-VAE and $\beta$-TCVAE, expect for batch-normalization.

\begin{figure}[!htbp]
	\centering
	\includegraphics[scale=1.3]{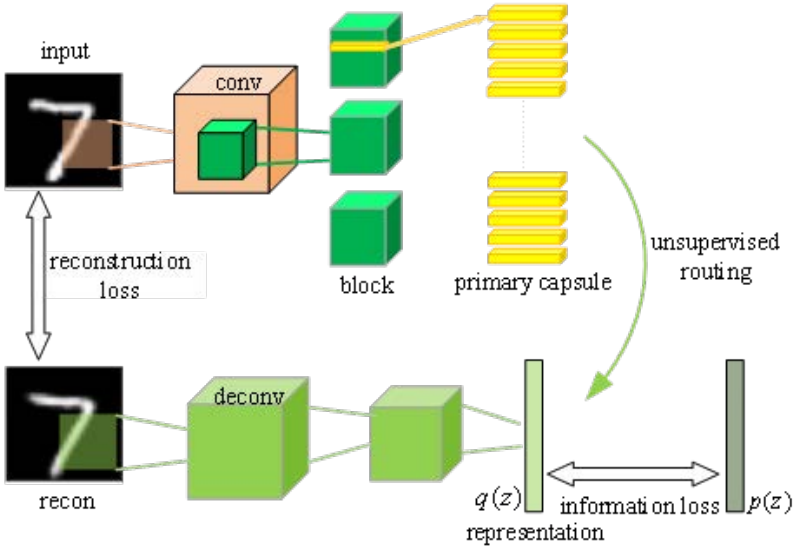}
	\caption{The structure of $\beta$-CapsNet in an unsupervised manner}
	\label{fig8}
\end{figure}

\begin{table}[!htbp]%表 multirow合并行  multicolumn合并列
	\centering
	\label{tb2}
	\caption{Structure of $\beta$-CapsNet in unsupervised learning}
	\begin{tabular}{cc}
		\hline
		%\multirow{2}{*}{Dataset} & \multirow{2}{*}{C} & \multicolumn{3}{c}{Sequence   Length} & \multirow{2}{*}{Events} \\ \cline{3-5}
	      \multicolumn{2}{c}{Unsupervised $\beta$-CapsNet} \\ \hline
input	 &(3, 64, 64)\\
encoder	&Conv (32, 4×4, 2)\\
    &Conv (64, 4×4, 2)\\
&Conv (128, 4×4, 2)\\
&Conv (64, 4×4, 1)\\
primary capsule &(576, 8)\\
routing matrix	&(576, 8, 16)\\
  & \\
representation	&(1,16)\\
decoder	&Deconv (512, 1×1, 1)\\
&Deconv (64, 4×4, 1)\\
&Deconv (64, 4×4, 2)\\
&Deconv (32,4×4, 2)\\
&Deconv (32,4×4, 2)\\
&Deconv (3,4×4, 2)\\
output	&(3, 64, 64) \\ \hline
	\end{tabular}
\end{table}

\section{Experimental results}

In this section, we would discuss $\beta$-CapsNet for supervised learning and unsupervised learning respectively. All the models were implemented using Pytorch and RTX-2070. For the training procedure, we used Adam optimizer with an initial learning rate of 0.001 and all the models are trained for 100 epochs.

In the first group of experiments, we carry out several experiments to validate information bottleneck loss in a supervised manner, we would perform a series of quantitative and qualitative experiments, showing that the relationship between reconstruction fidelity, classfication task and the quality of disentanglement by the trade-off parameter $\beta$. We analyze the classification performance and upper bound on the loss of our proposed $\beta$-CapsNet and CapsNet with different values of $\beta$ and datasets. Then we compare the effects of different $\beta$ on the reconstructions of the decoder, we find the configuration for the best performance about the dimension of the representation and the value of $\beta$. Finally, we train $\beta$-CapsNet with appropriate hyperparameters on two datasets commonly used to evaluate disentangling performance on supervised learning. 

In the second group of experiments, we validate the effectiveness of information bottleneck loss for an unsupervised manner and confirm qualitatively that our model discovers more disentangled factors than CapsNet and $\beta$-VAE while also being fairly robust to random initialization on unsupervised MNIST \& Fashion-MNIST, 3D chairs and CelebA datasets.

\subsection{$\beta$-CapsNet for Supervised Data}

\subsubsection{Supervised Datasets}

(1)\textbf{MNIST} \cite{36lecun1998mnist}: Modified national institute of standards and technology database is a basic dataset of handwritten digits that is commonly used for computer vision task, it contains 60k training images and 10k testing images, each of them is a 28×28 gray image. There is some interpretable semantic information between images of same class, therefore it is one of the supervised data that is often used to verify disentanglement.

(2)\textbf{Fashion-MNIST} \cite{37xiao2017fashion}: As a replacement and strengthening benchmarking dataset for the original MNIST, Fashion-MNIST is a dataset of article images consisting of the same amount and size. It also contains several separate factores of variation in the data of same class, so we adopt it as another supervised dataset.

\subsubsection{Disentanglement Trade-off}

The parameter $\beta$ in our method is used to adjust the information bottleneck loss of the representation, it can be seen as managing the tradeoff among the disentanglement of the represetation (measured by information loss), the fidelity of the reconstruction of the input from the representation (reconstruction loss) and classification accurucy (margin loss). In this subsection, we would compare our method with CapsNet baseline \cite{8sabour2017dynamic} on some standard benchmark using different values of $\beta$.

\textbf{Classification Performances of Variation $\beta$:}Different from unsupervised learning scene, the input constains some test samples in supervised learning which are not involved in the training process, since there is a difference in classifiaction performances between training and test set. Here we only need to demonstrate the accuracies of the test set, because the accuracies of the training set is always unconsidered. We set $\beta  \in (0,\;0.01,\;0.1,\;1,\;2,\;3,\;4,\;5)$, when $\beta  = 0$ we get back the original CapsNet. 

The classifiaction performances are depicted in Fig.9 that confirms our intuition: when training with small values of $\beta$, the network has very little pressure to limit the information of the representation, the classified capsules have enough information to finish precise classification, so we can expect our model to achieve better performance; on the other hand, increasing the value of $\beta$ make a more strongly information constraint on the representation, therefore the model tends to find more disentangled factors rather than other tasks during the training, we expect the degradation in performance. 

\begin{figure}[!htbp]
	\centering
	\includegraphics[scale=0.5]{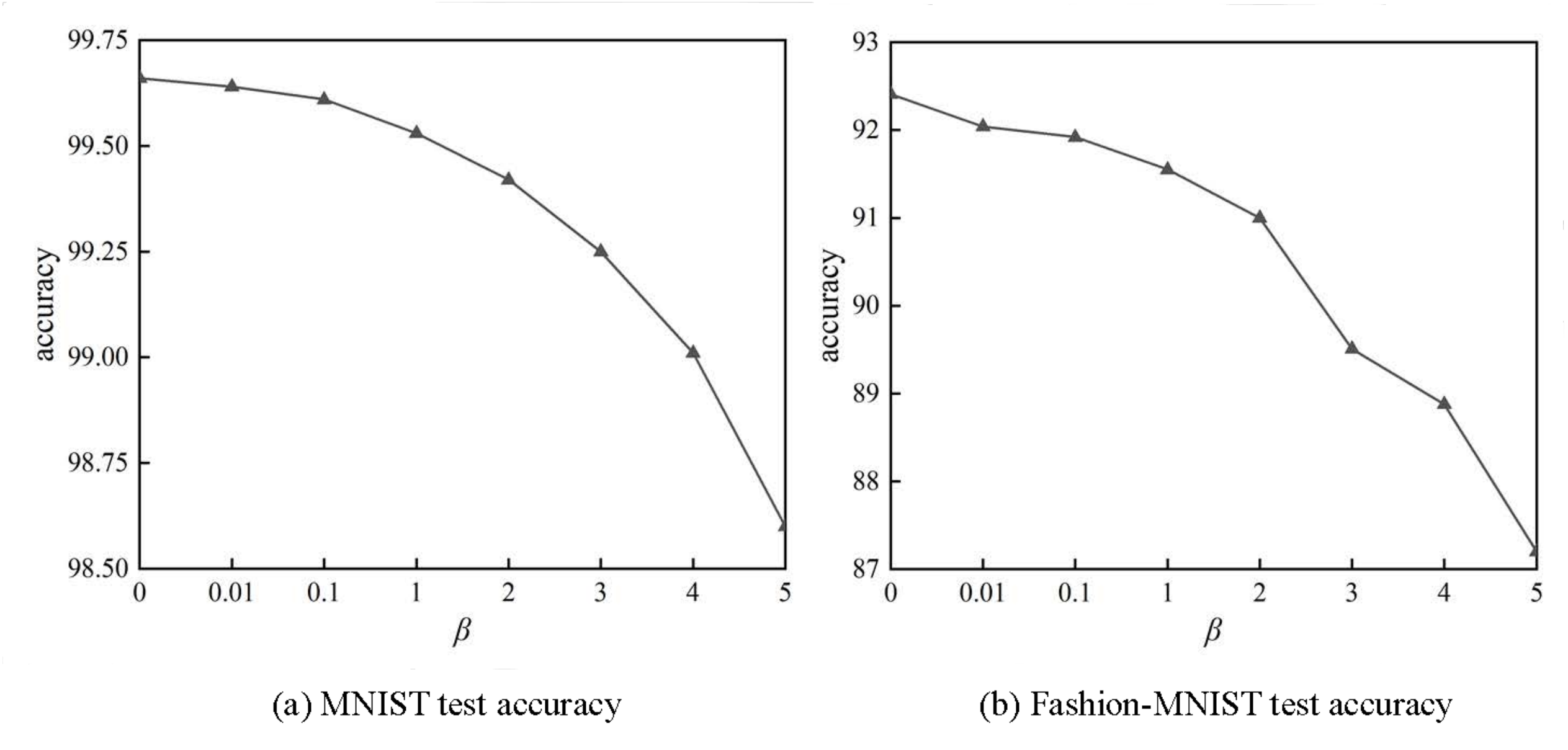}
	\caption{ Classification accuracies on MNIST (a) and Fashion MNIST (b). When $\beta$=0.01, the accuracy of our method is close to CapsNet, and the accuracy decreases gradually when the weight $\beta$ increases, this process indicates that this kind of information constraint will affect the performance of classification task.}
	\label{fig9}
\end{figure}

\textbf{Variational Loss of Variation $\beta$:}Since the $\beta$-CapsNet’s loss is upper bounds on the standard loss function, we would like to see the effect of changing the value of $\beta$ on training and test set. We train several $\beta$-CapsNet using a group of different values $\beta  \in (0,\;0.01,\;0.1,\;1,\;3,\;5,\;8,\;10)$, when $\beta  = 0$ we get back the original CapsNet (its information loss is none). To paint a clearer picture, we aggregate total loss, information loss, reconstruction loss and margin loss to visualize the effect of the parameter $\beta$ in training set as shown in Fig.9 (a) and (b), then we visualize total loss and reconstruction loss with variation in Fig.9 (c) and (d) because the margin loss and information loss in the test set have almost the same curves as the training set. 

\begin{figure}[htbp]%多图片
\centering
\subfigure[The MNIST training set                                           \quad \quad \quad \quad \quad  (b) The Fashion-MNIST training set]
{
    \begin{minipage}[b]{.8\linewidth}
        \centering
        \includegraphics[scale=1]{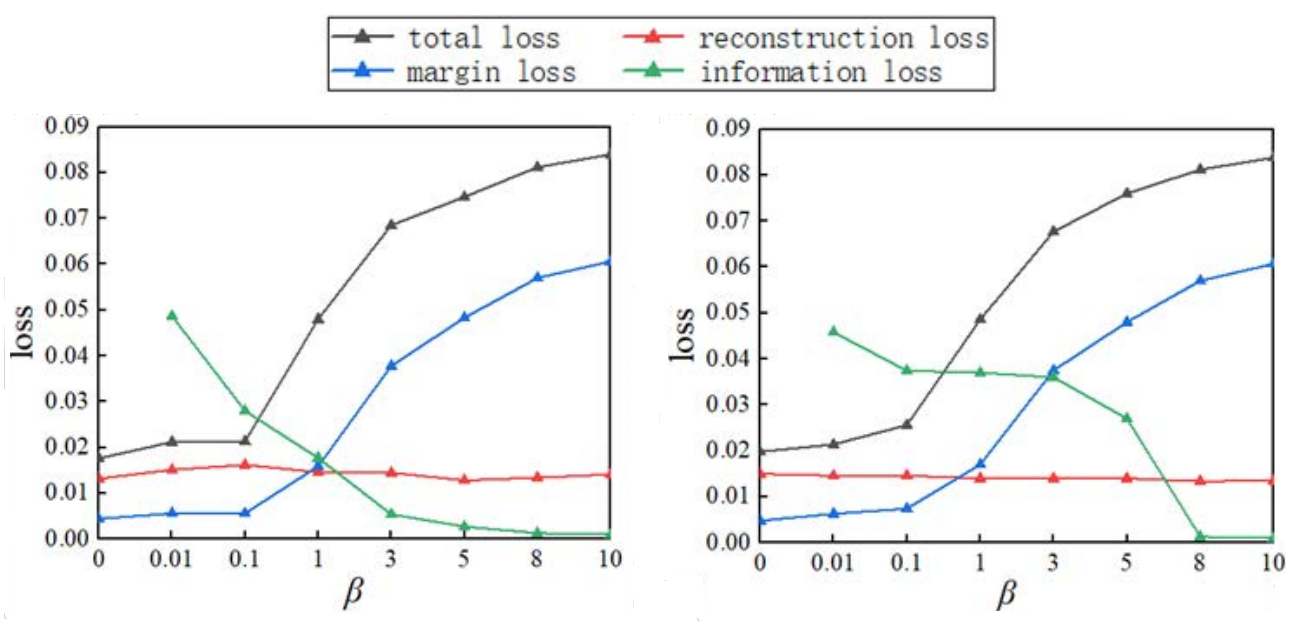}
    \end{minipage}
}
\subfigure
{
}
\subfigure[The MNIST test set                                                     \quad \quad \quad \quad \quad (d) The Fashion-MNIST test set ]
{
 	\begin{minipage}[b]{.8\linewidth}
        \centering
        \includegraphics[scale=1]{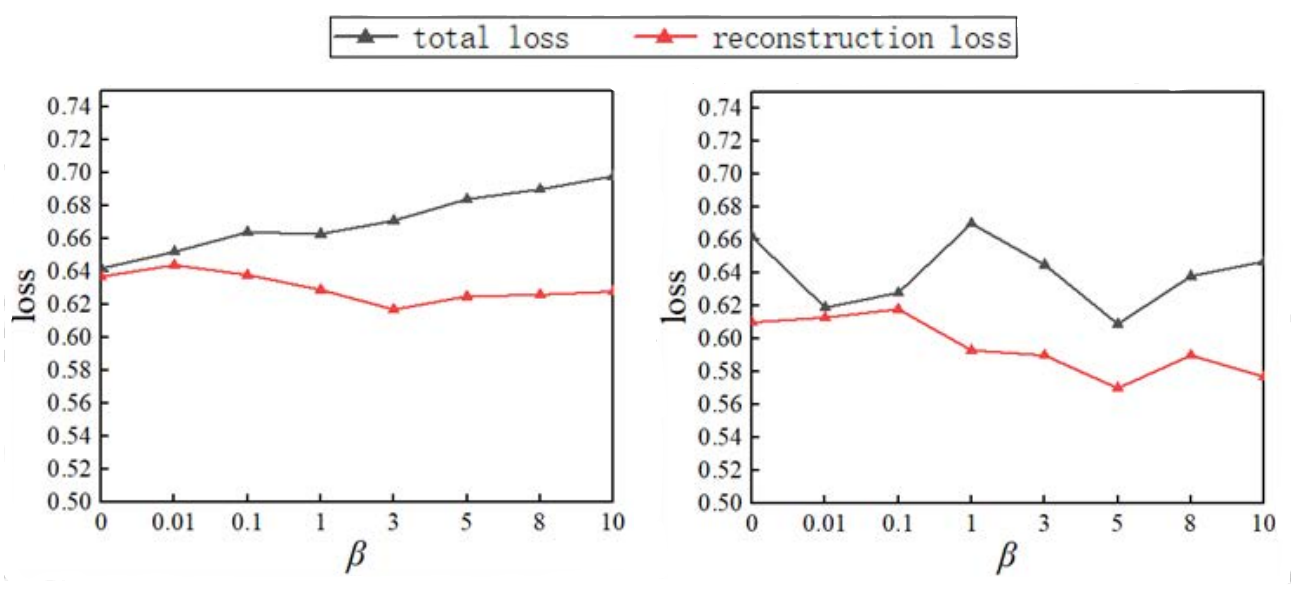}
    \end{minipage}
}
\caption{When training with small values of $\beta$, weak information constraint on the representation leads to a high information loss, small margin loss corresponds to better classification performance. In test set, the reconstruction loss accounts for the most part in total loss, it is 30 times that of the training set. As $\beta$ increases, reconstruction loss tends to decrease for Fashion-MNIST dataset but there is no obvious fluctuation for MNIST dataset.}
\end{figure}

Maigin loss and information loss are in line with our expectations: when training with large values of $\beta$, the network attends to limit the information of the representation, severe constraint on the representation leads to small information loss, meanwhile, margin loss increases due to small restriction corresponding to the decline in accuracies. Although in theory the increase of $\beta$ will increase reconstruction loss that in turn blurs reconstructed image, we observe that reconstructions effects in Fig.10 (a), (b) and (c) are almost unaffected. Therefore, we would to explore the influence of $\beta$ on reconstructed images from Fig.11 in the next subsection, the experimental results show that the reconstructed samples of $\beta$-CapsNet are almost unaffected by our information bottleneck constraints, this is remakable different from $\beta$-VAE whose reconstruction image details will be blurred seriously as the constraints increase.

\begin{figure}[!htbp]
	\centering
	\includegraphics[scale=1]{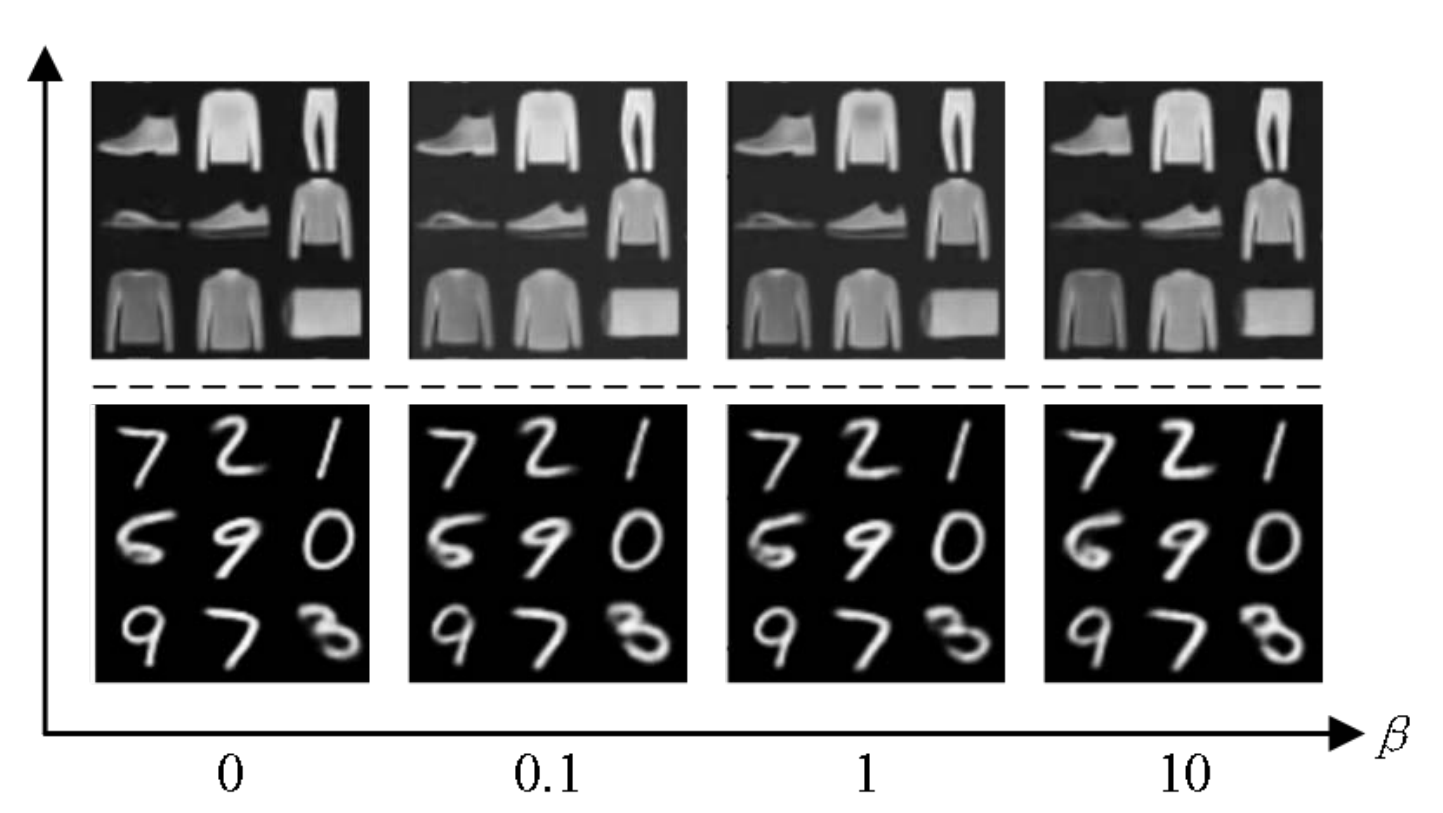}
	\caption{ When $\beta$ is set to 0, the $\beta$-CapsNet reduces to the conventional CapsNet, its reconstructions are slightly sharper than the reconstructions of the other three models, no matter which dataset is, it shows that the information constraints on the representation has little impact on the reconstructed images of the decoder, which may depend on the strong expression ability of the capsule vector in the representation. The other three groups of reconstructions have almost the same sharpness, regardless of the value of $\beta$, this result is consistent with the reconstruction loss, which is one of the noteworthy advantages of $\beta$-CapsNet.}
	\label{fig11}
\end{figure}

\subsubsection{Qualitative Comparisons of Disentanglement}

In order to qualitatively compare the disentangling performance of $\beta$-CapsNet against CapsNet on supervised dataset, we train these models on MNIST and Fashion-MNIST. The components of capsule for CapsNet are set within the range of [-0.2, 0.2], the components of capsule for $\beta$-CapsNet are set within the range of [-0.08, 0.08] and $\beta$ is set to 3. Fig.12 depicts interpretable properties in representation: both $\beta$-CapsNet and CapsNet have shown to be capable of learning several properties including thickness, width and angle on MNIST, width and length on Fashion-MNIST. However, CapsNet always tends to learn entangled factors, for instance, digit thickness in Fig.12 (a) is entangled with angle, cloth length in Fig.12 (b) is entangled with width, CapsNet can only perceive the width variation in two categories. In contrast, $\beta$-CapsNet learns more disentanglement factors which are more interpretable.

%\begin{table}[!htbp]%表 multirow合并行  multicolumn合并列
%	\centering
%	\begin{tabular}{cc}
%		%\multirow{2}{*}{Dataset} & \multirow{2}{*}{C} & \multicolumn{3}{c}{Sequence   Length} & \multirow{2}{*}{Events} \\ \cline{3-5}
%	CapsNet  &     $\beta$-CapsNet\\
%\includegraphics[scale=1]{Object 87.pdf} &\includegraphics[scale=1]{Object 88.pdf}\\
%\multicolumn{2}{c}{Thickness}\\
%\includegraphics[scale=1]{Object 89.pdf} &\includegraphics[scale=1]{Object 90.pdf}\\
%\multicolumn{2}{c}{Width}\\
%\includegraphics[scale=1]{Object 91.pdf} &\includegraphics[scale=1]{Object 92.pdf}\\
%\multicolumn{2}{c}{Angle}\\
%\multicolumn{2}{c}{(a) Models trained on MNIST}\\
%CapsNet	&$\beta$-CapsNet\\
%\includegraphics[scale=1]{Object 93.pdf} &\includegraphics[scale=1]{Object 94.pdf}\\
%\multicolumn{2}{c}{Width}\\
%\includegraphics[scale=1]{Object 95.pdf} &\includegraphics[scale=1]{Object 96.pdf}\\
%\multicolumn{2}{c}{Length}\\
%\multicolumn{2}{c}{(b) Models trained on Fashion-MNIST}\\
%\multicolumn{2}{c}{Figure 12: Qualitative comparing disentangling results of CapsNet and $\beta$-CapsNet}\\
%	\end{tabular}
%\end{table}

\begin{figure}[!htbp]
	\centering
	\includegraphics[scale=0.97]{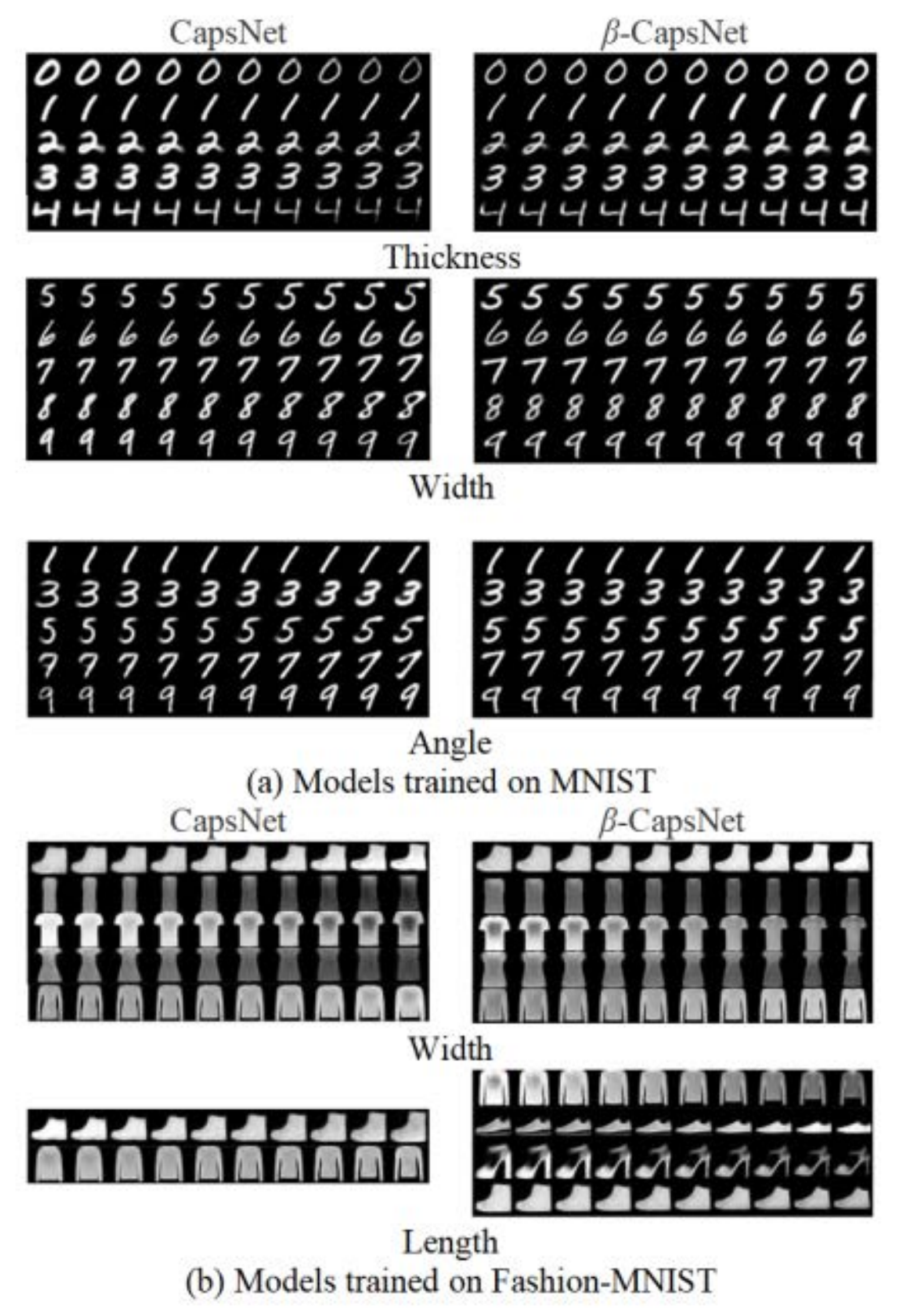}
	\caption{Qualitative comparing disentangling results of CapsNet and $\beta$-CapsNet}
	\label{fig12}
\end{figure}

\subsection{$\beta$-CapsNet for Unsupervised Data}

\subsubsection{Unsupervised Datasets}

(1) \textbf{Unsupervised MNIST \& Fashion-MNIST:} Handwritten digits and clothes without labels are approprate unsupervised datasets for learning disentangle factors, we would like to observe interpretability of capsule and different disentangled factors from unsupervised learning viewpoint.

(2) \textbf{3D Chairs \cite{38aubry2014seeing}:} 3D Chairs is a large dataset of many chair classes using for object category detection in images as a type of 2D to 3D alignment problem. A chair class can be seem as a running example which contains several continuous interpretable features, therefore it has become a dominant dataset for disentangled representation evaluation.

(3) \textbf{CelebA \cite{39liu2015deep}:} CelebFaces attributes dataset is a dataset for face attributes with more than 10k number of identities and 200k face images, it can be employed as the training and test sets for many computer vision tasks that is one of the most dominant dataset for learning disentangled representation. As a general preprocessing step, the aligned images are center cropped to 128×128 and then downsampled to 64×64, the center crop can remove background and make it easier for reconstructions.

\subsubsection{Qualitative Comparisons on Unsupervised MNIST}

The differences of category number in MNIST dataset without label are the most significant interpretable factor. Therefore, in addition to factors within the same category, disentangled learning should be able to learn to control transformations between categories of similar shapes such as 7 to 9. Fig.13 provides a qualitative comparison of the disentangling performance of CapsNet and $\beta$-CapsNet, The components of capsule for CapsNet are set within the range of [-0.3, 0.3] or [-0.6, 0.6] ($\beta$ is set to 0), the components of capsule for $\beta$-CapsNet are set within the range of [-0.15, 0.15] and $\beta$ is set to 0.2. It can be seen that both models are able to automatically identify and learn the disentangle factors such as thickness, width and angle, however, $\beta$-CapsNet can consistently and significantly learn more disentangled latent capsule which is more obvious than the disentangling performance on supervised dataset in Fig.9 (a). For example, when learning about thickness factor, CapsNet without information bottleneck constraint entangles digit width or angle with thickness in almost all digit classes.

Further, although CapsNet performs relatively well on transformations between part of the digital category, it still struggles to learn a clean factor between multiple categories. By contrast, $\beta$-CapsNet can learn a variety of transformations between two categories and multiple categories covering almost all digital categories of similar shapers. This experimental result suggests that a discrete disentanglement quality can be controlled by continuing latent representation of CapsNet, and the information bottleneck constraint leads to better disentanglement both in the same category and in different categories.

\begin{figure}[!htbp]
	\centering
	\includegraphics[scale=0.8]{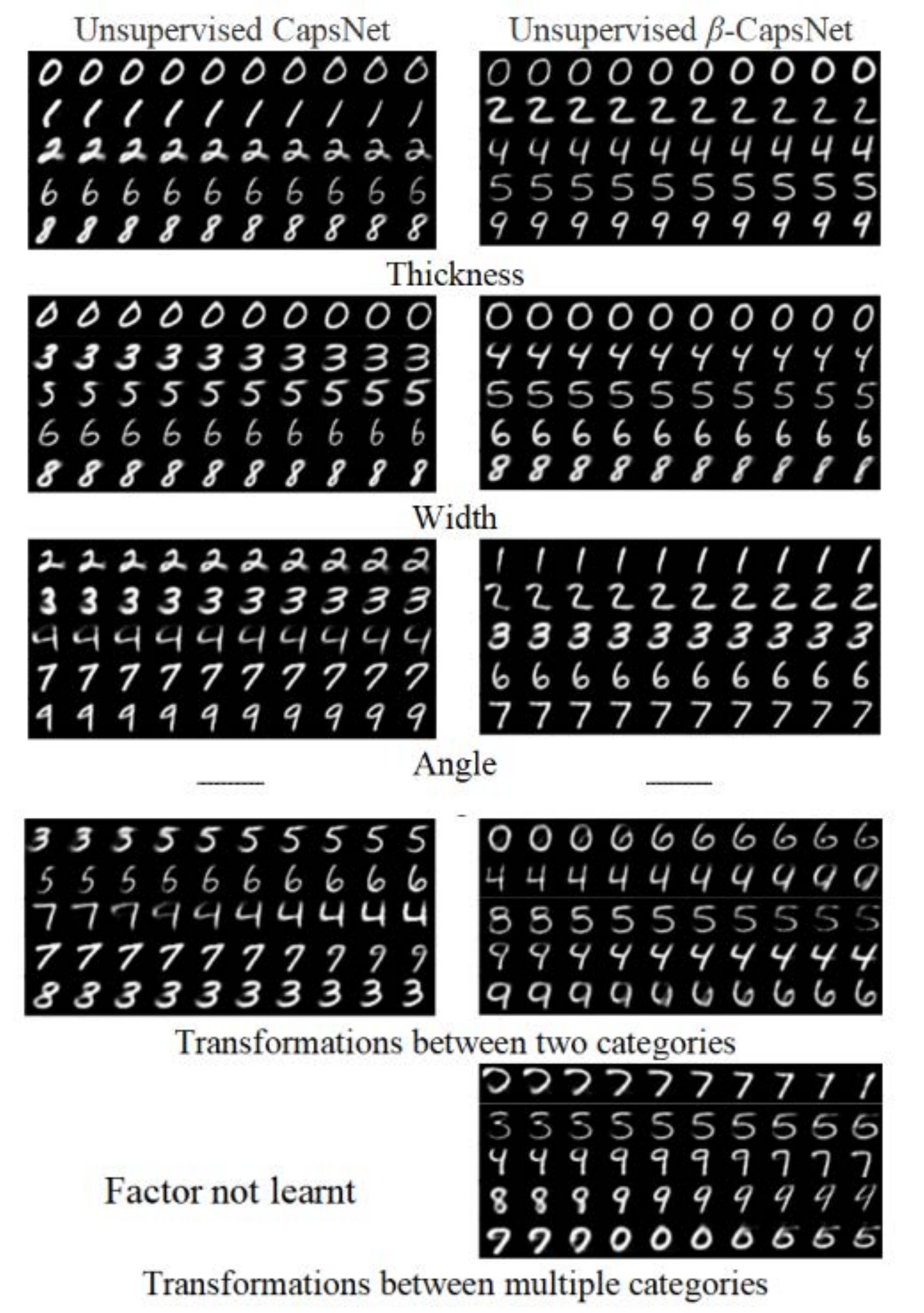}
	\caption{Qualitative comparing results of disentanglement for unsupervised CapsNet and $\beta$-CapsNet}
	\label{fig13}
\end{figure}

\subsubsection{Qualitative Comparisons on 3D Chairs}

Manipulating latent variables on 3D chairs are often used for comparing qualitative results of disentangling performance, Fig.14 depicts the interpretable properties in reconstructing 3D chairs from latent representation of $\beta$-VAE \cite{4bengio2013representation}, $\beta$-TCVAE \cite{14chen2018isolating} and $\beta$-CapsNet. However, most properties learned by $\beta$-VAE and $\beta$-TCVAE are entangled with others, for instance, chair size is entangled with chair category, backrest is entangled with azimuth and chair category. By contrast, the representation learned by $\beta$-CapsNet is disentangled with nuances. $\beta$-TCVAE and $\beta$-CapsNet are capable of learning an additional property: rotation for swivel chairs, this property is more subtle and likely require a higher mutual information (total correlation mutual information in $\beta$-TCVAE). The shortcoming of our model is that azimuth learned by $\beta$-CapsNet is not as good as $\beta$-TCVAE.

\begin{figure}[htbp]%多图片
\centering
\subfigure
{
    \begin{minipage}[b]{.8\linewidth}
        \centering
        \includegraphics[scale=0.9]{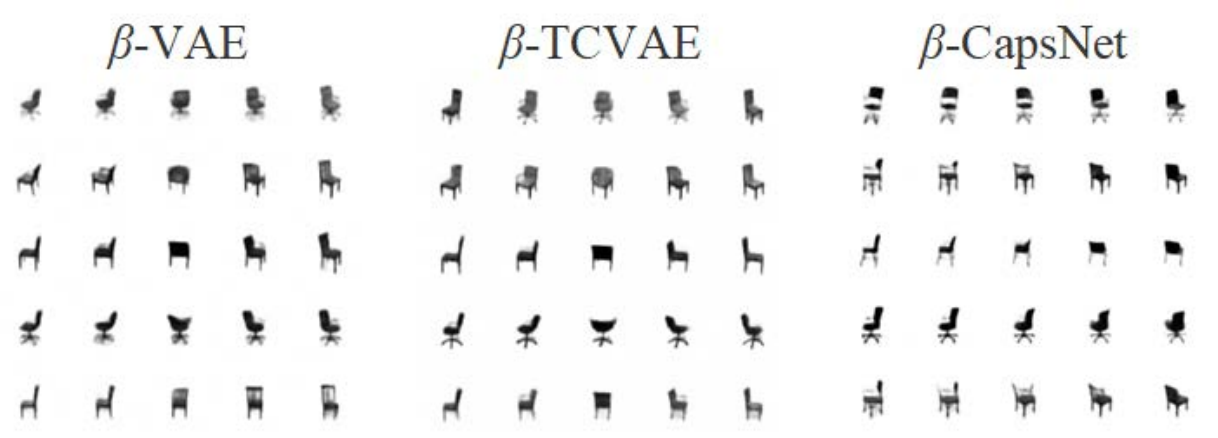}
    \end{minipage}
}
\subfigure
{
 	\begin{minipage}[b]{.8\linewidth}
        \centering
        \includegraphics[scale=0.9]{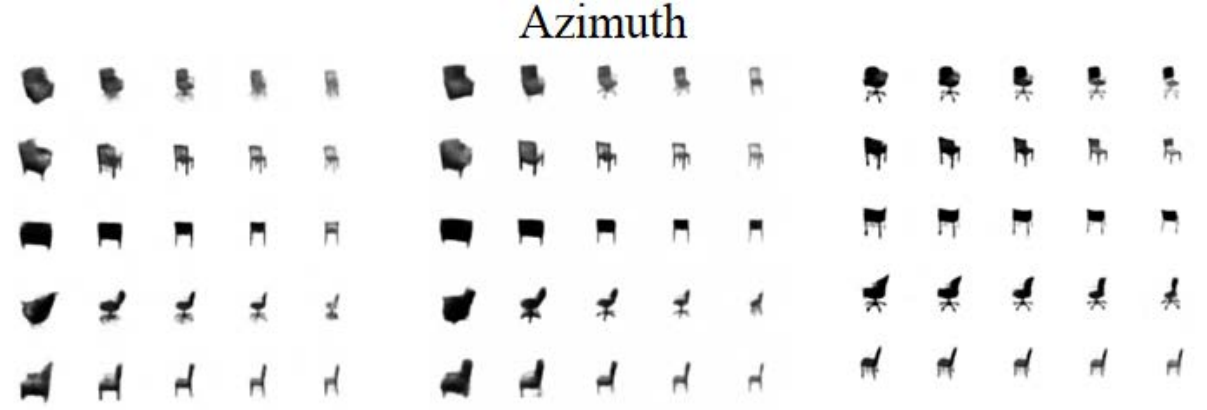}
    \end{minipage}
}
\subfigure
{
 	\begin{minipage}[b]{.8\linewidth}
        \centering
        \includegraphics[scale=0.9]{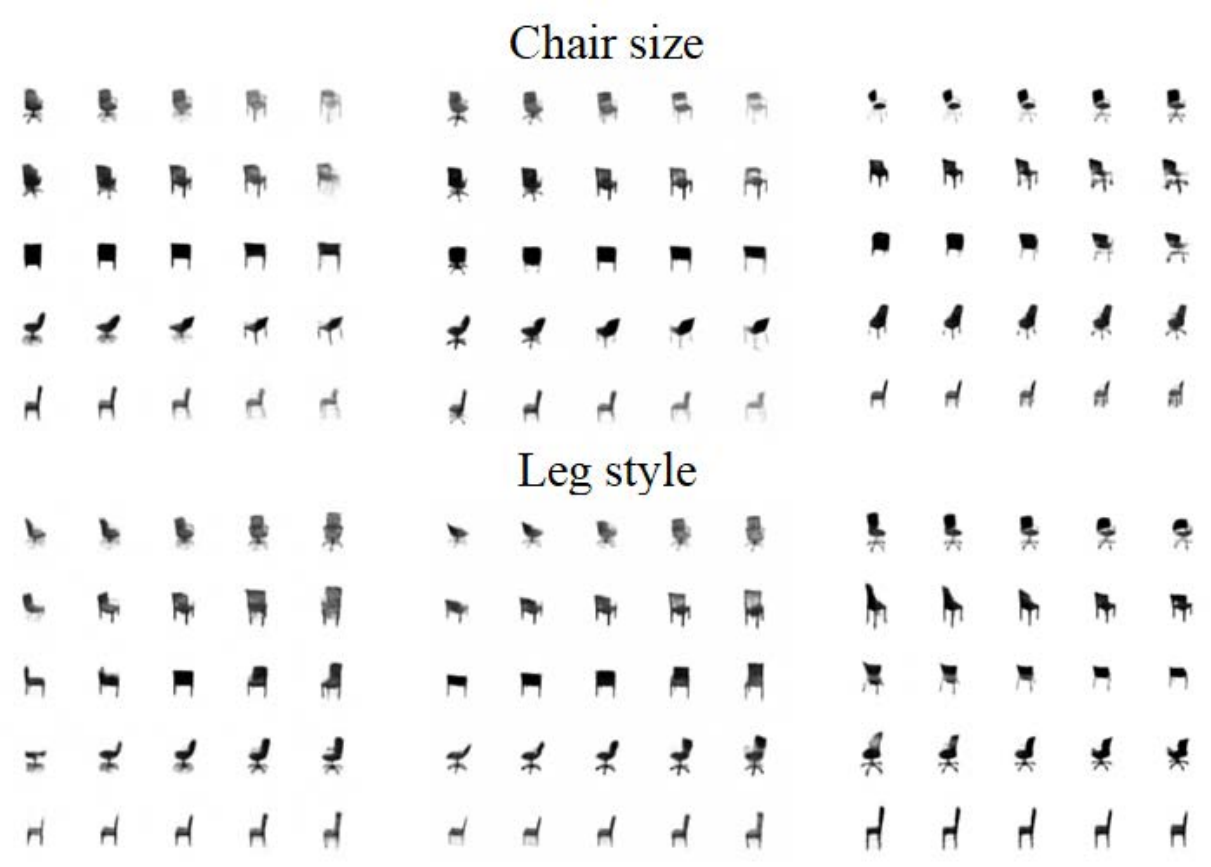}
    \end{minipage}
}
\subfigure
{
 	\begin{minipage}[b]{.8\linewidth}
        \centering
        \includegraphics[scale=0.9]{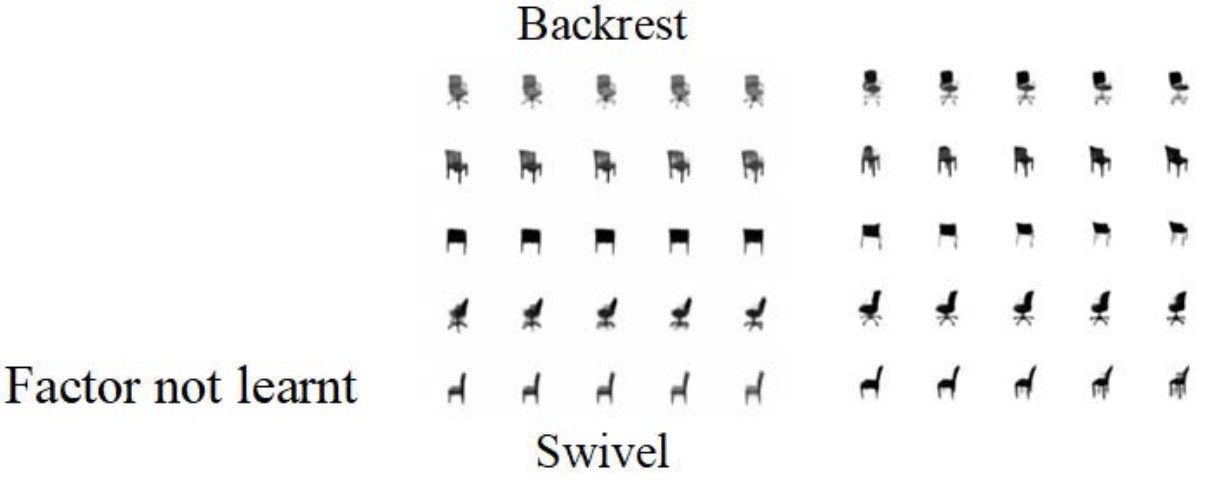}
    \end{minipage}
}
\caption{Disentanglement results of unsupervised $\beta$-CapsNet on 3D chair}
\end{figure}

\subsubsection{Disentangled Capsule on Fashion-MNIST and CelebA}

\textbf{Unsupervised Fashion-MNIST:} Fig.15 shows that width and length attributes in 4 classes are discovered by unsupervised $\beta$-CapsNet, these attributes are the same as the disentangled latent factors with supervision in Fig.12 (b). Furthermore, unsupervised $\beta$-CapsNet can learn some transformation factors of capsule between two categories (e.g. trouser to pullover) and multiple categories (e.g. sandal to sneaker to coat), supervised $\beta$-CapsNet leads to entangled capsule in different categories due to the mask vector.

\begin{figure}[htbp]%多图片
\centering
\subfigure[Width]
{
    \begin{minipage}[b]{.8\linewidth}
        \centering
        \includegraphics[scale=1]{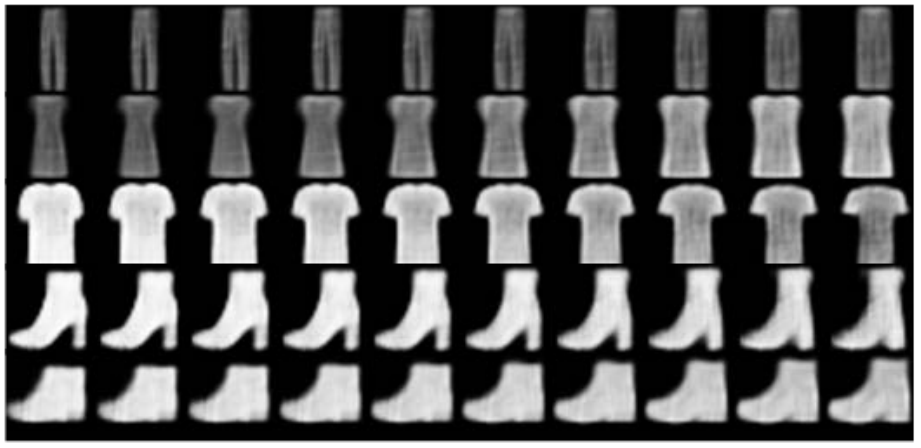}
    \end{minipage}
}
\subfigure[Length]
{
 	\begin{minipage}[b]{.8\linewidth}
        \centering
        \includegraphics[scale=1]{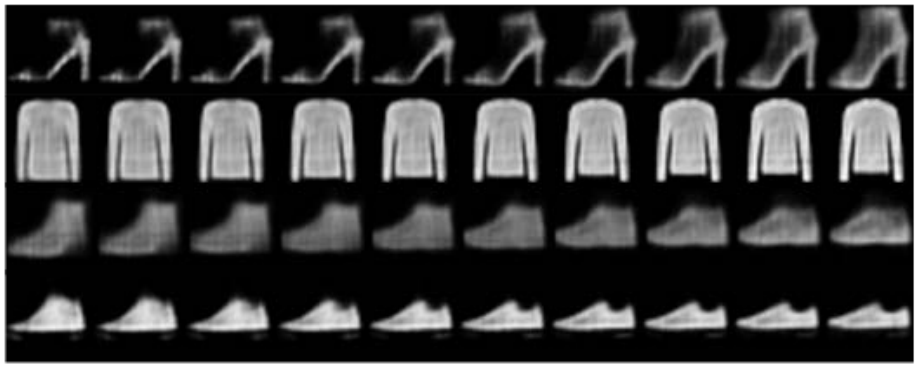}
    \end{minipage}
}
\subfigure[Transformations between different categories]
{
 	\begin{minipage}[b]{.8\linewidth}
        \centering
        \includegraphics[scale=1]{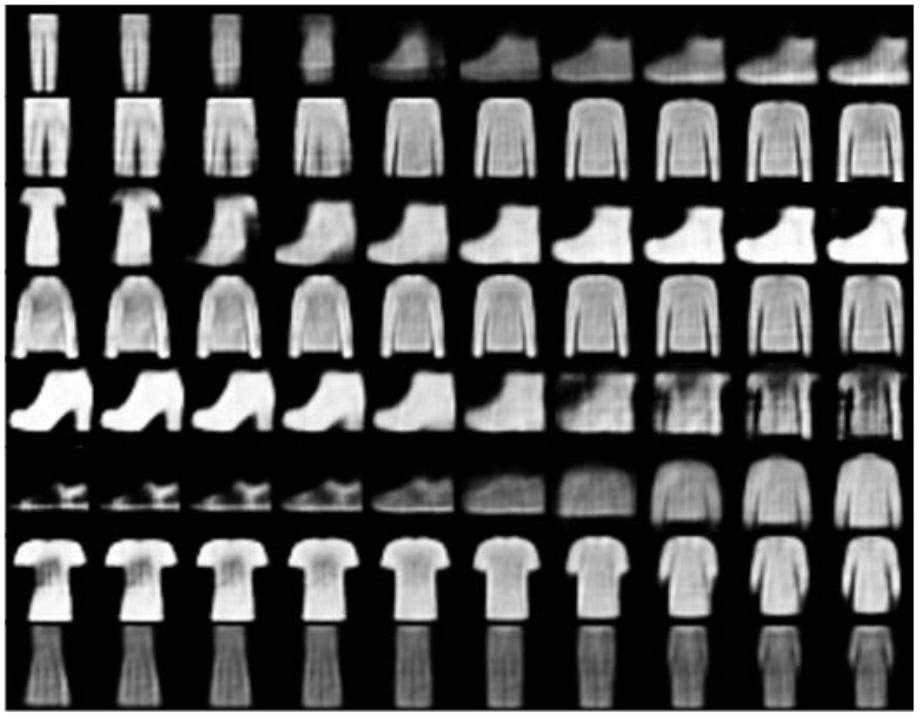}
    \end{minipage}
}
\caption{Disentanglement results of unsupervised $\beta$-CapsNet on Fashion-MNIST}
\end{figure}

\textbf{CelebA:} Fig.16 shows that 11 attributes out of 16 dimension are discovered by the $\beta$-CapsNet ($\beta$=1) without supervision. $\beta$-VAE discovers six disentangled factors only and some of them are entangled with nuances, $\beta$-CapsNet dose discover numerous extra factors such as bangs, masculinity and glasses. In addition, it is difficulty to render complete face width or skin color for $\beta$-VAE and $\beta$-TCVAE, whereas the experimental results for $\beta$-CapsNet show meaningful disentanglement and extrapolation characteristics. For instance, the extrapolation of face width for $\beta$-CapsNet shows that it focuses more on facial lines and contours, whereas the experimental results for $\beta$-TCVAE is entangled with many irrelevant factors such as azimuth and gender. 

\begin{figure}[htbp]%多图片
\centering
\subfigure[skin color]
{
    \begin{minipage}[b]{.3\linewidth}
        \centering
        \includegraphics[scale=0.3]{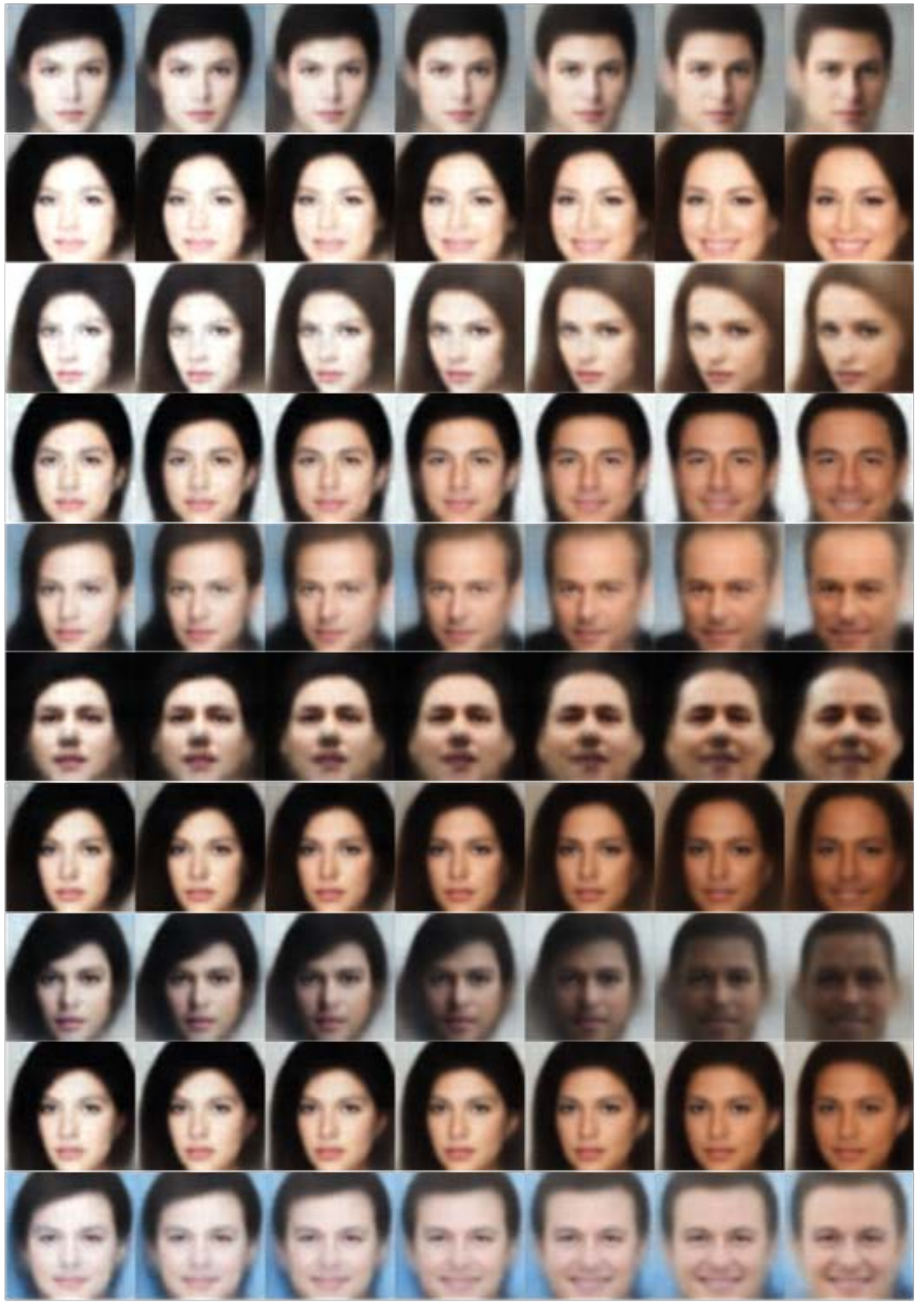}
    \end{minipage}
}
\subfigure[baldness]
{
 	\begin{minipage}[b]{.3\linewidth}
        \centering
        \includegraphics[scale=0.3]{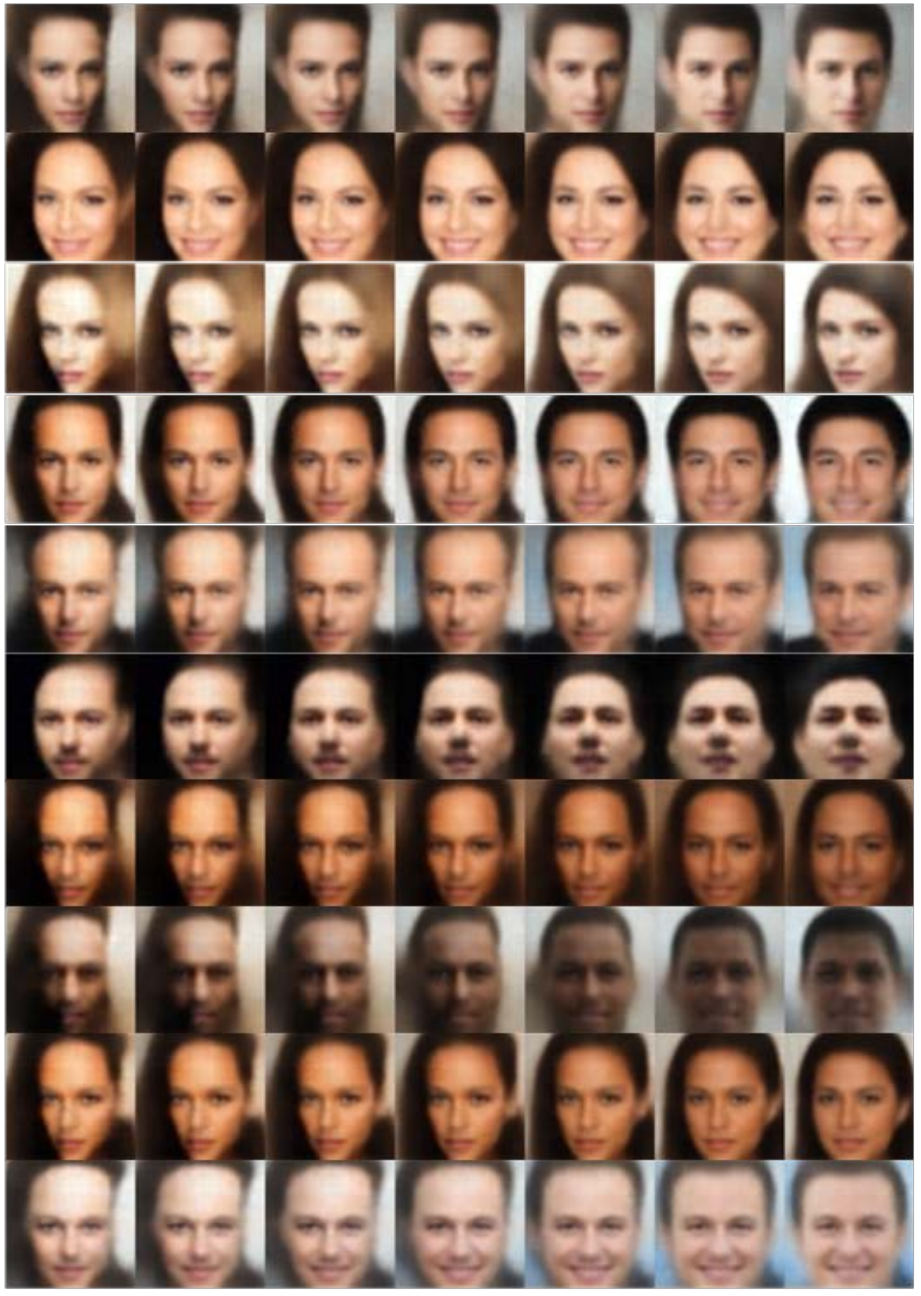}
    \end{minipage}
}
\subfigure[gender]
{
 	\begin{minipage}[b]{.3\linewidth}
        \centering
        \includegraphics[scale=0.3]{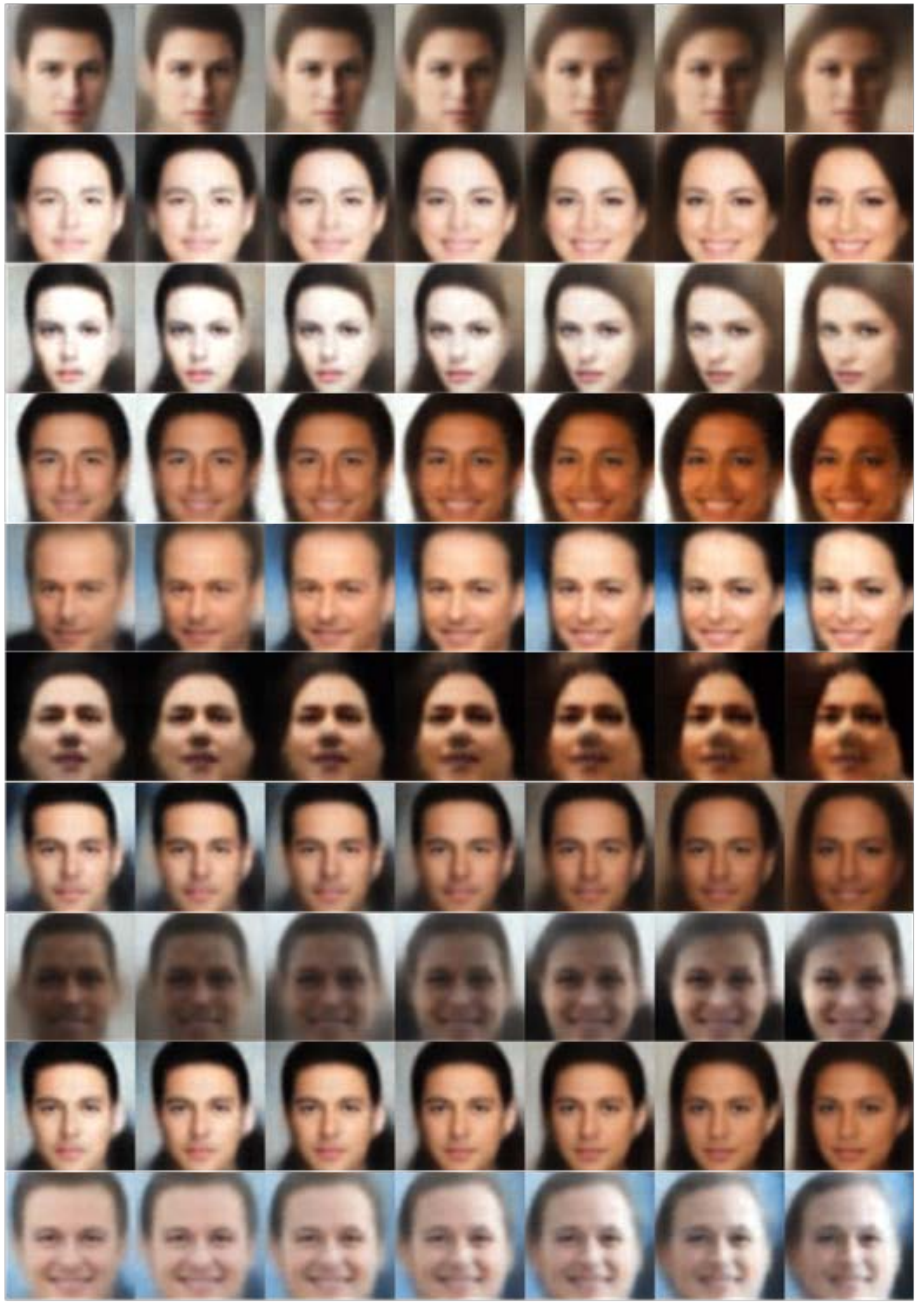}
    \end{minipage}
}
\subfigure[face width]
{
 	\begin{minipage}[b]{.3\linewidth}
        \centering
        \includegraphics[scale=0.3]{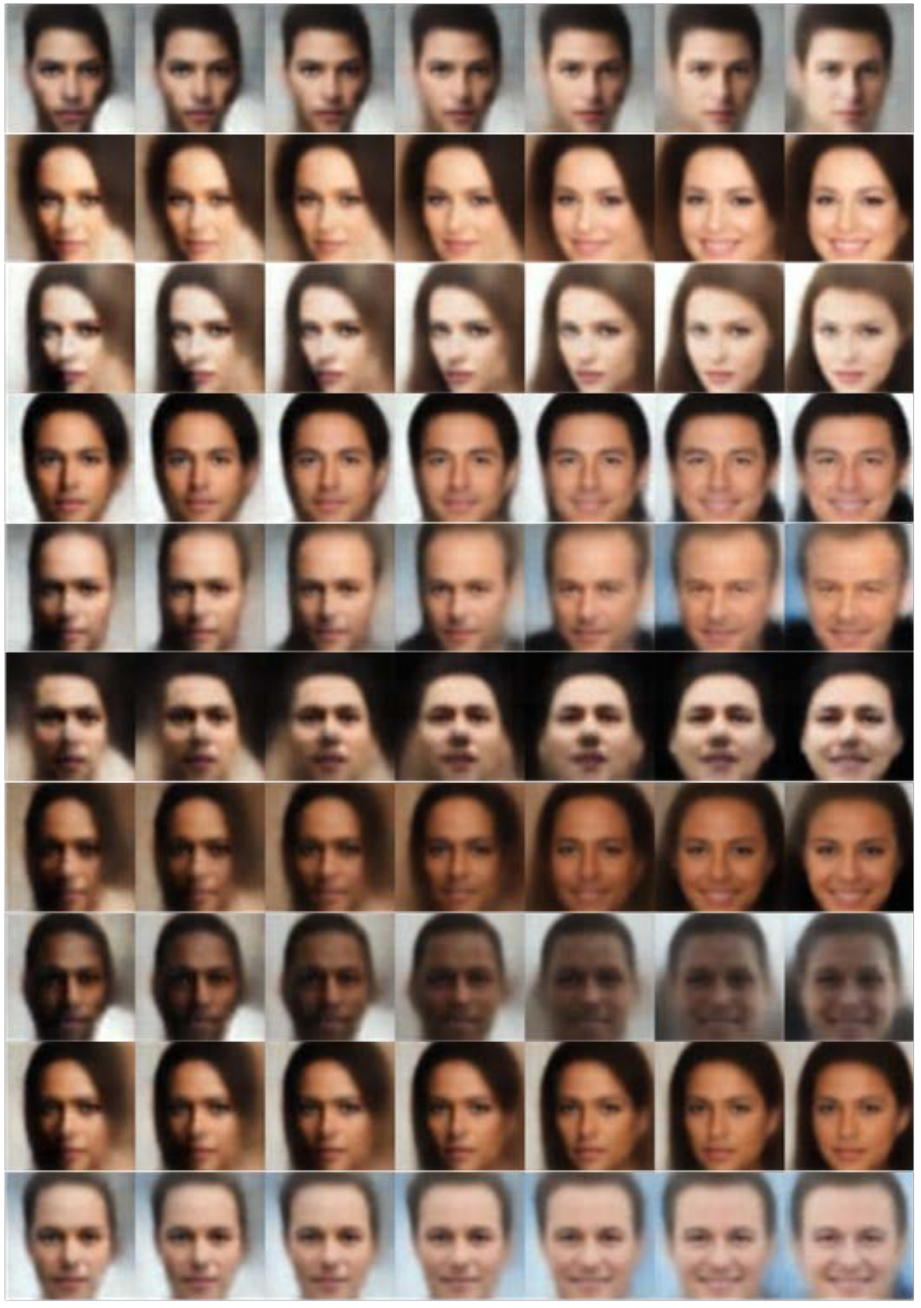}
    \end{minipage}
}
\subfigure[bangs]
{
 	\begin{minipage}[b]{.3\linewidth}
        \centering
        \includegraphics[scale=0.3]{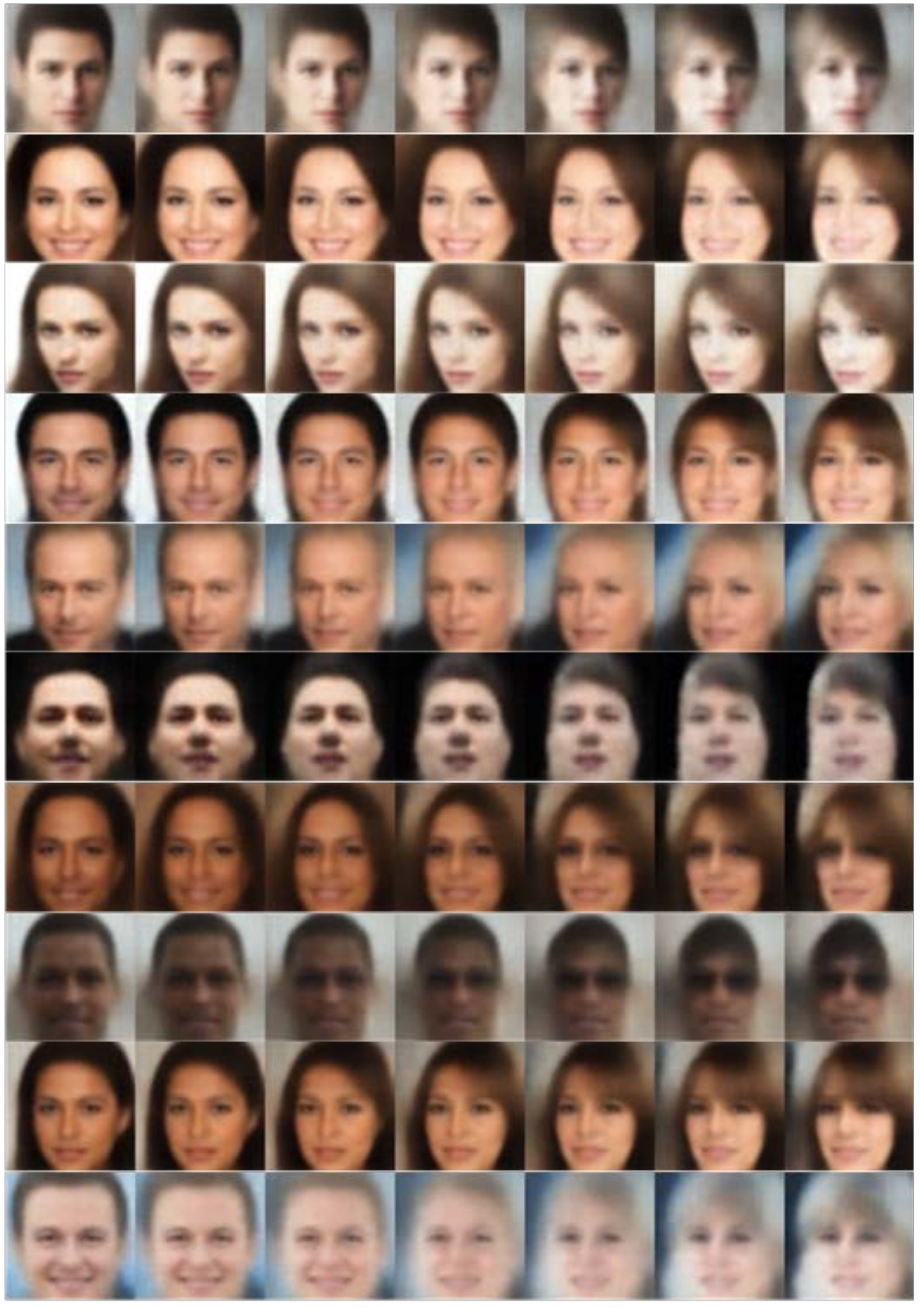}
    \end{minipage}
}
\subfigure[brightness]
{
 	\begin{minipage}[b]{.3\linewidth}
        \centering
        \includegraphics[scale=0.3]{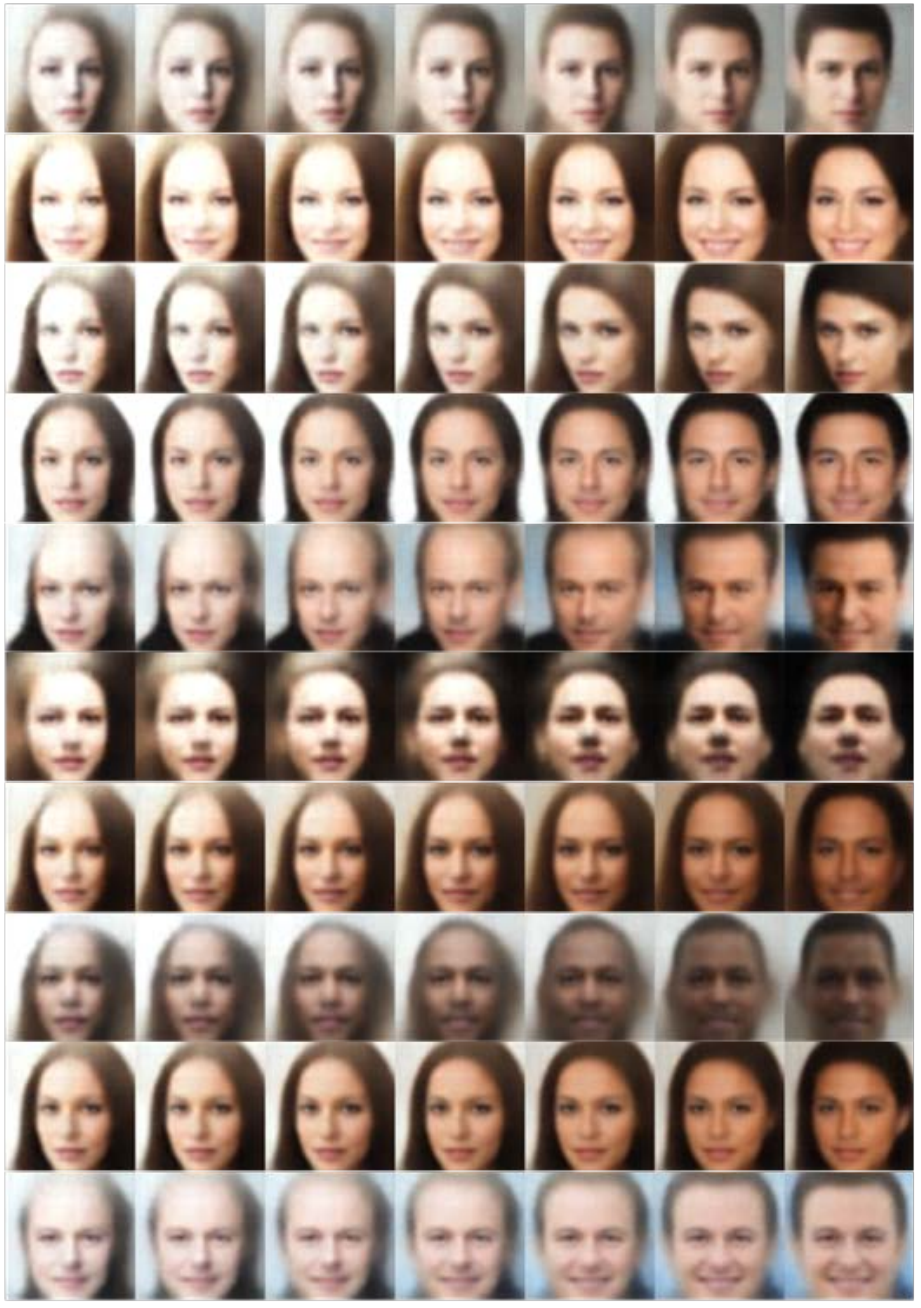}
    \end{minipage}
}
\subfigure[masculinity]
{
 	\begin{minipage}[b]{.3\linewidth}
        \centering
        \includegraphics[scale=0.3]{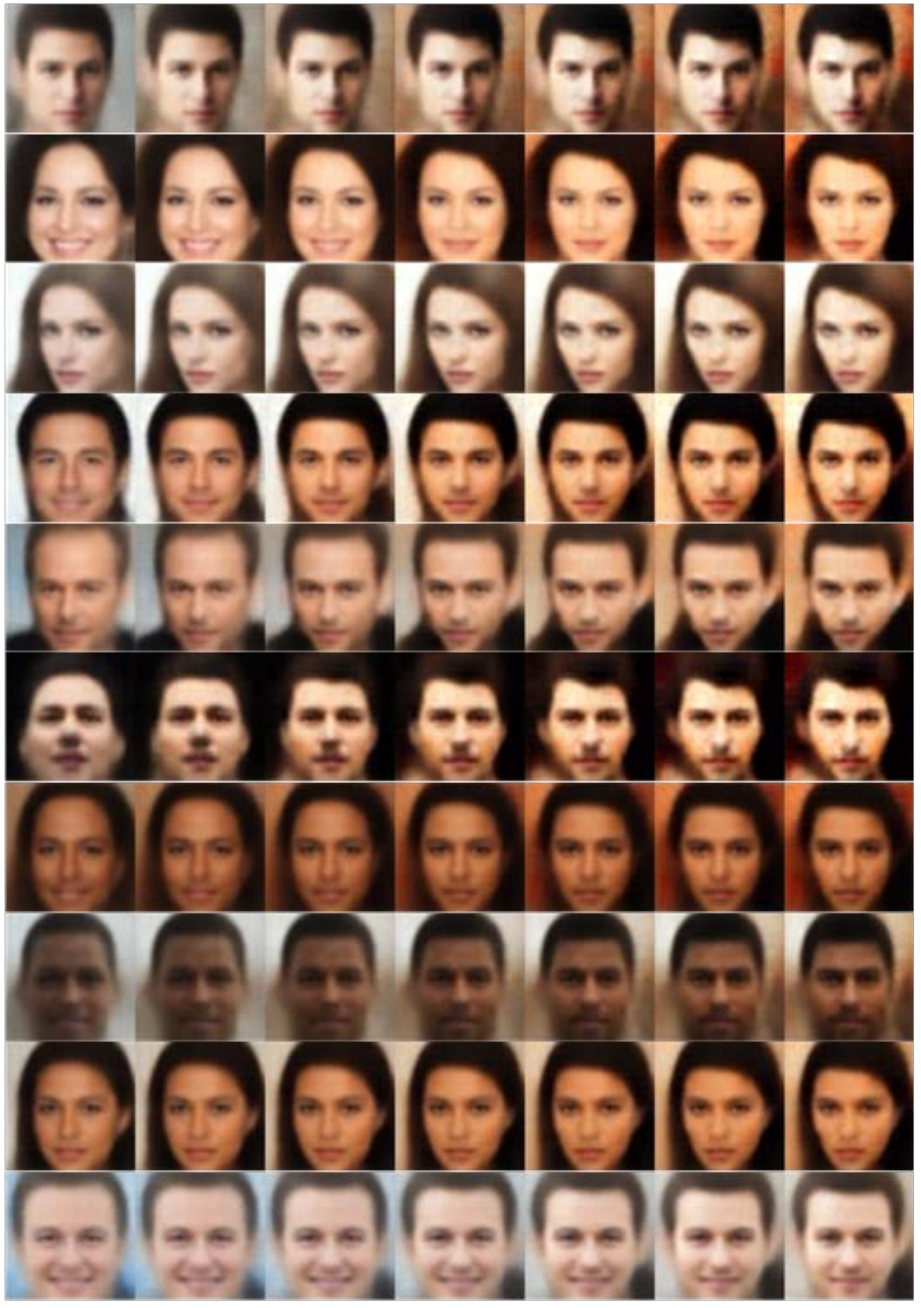}
    \end{minipage}
}
\subfigure[glasses]
{
 	\begin{minipage}[b]{.3\linewidth}
        \centering
        \includegraphics[scale=0.3]{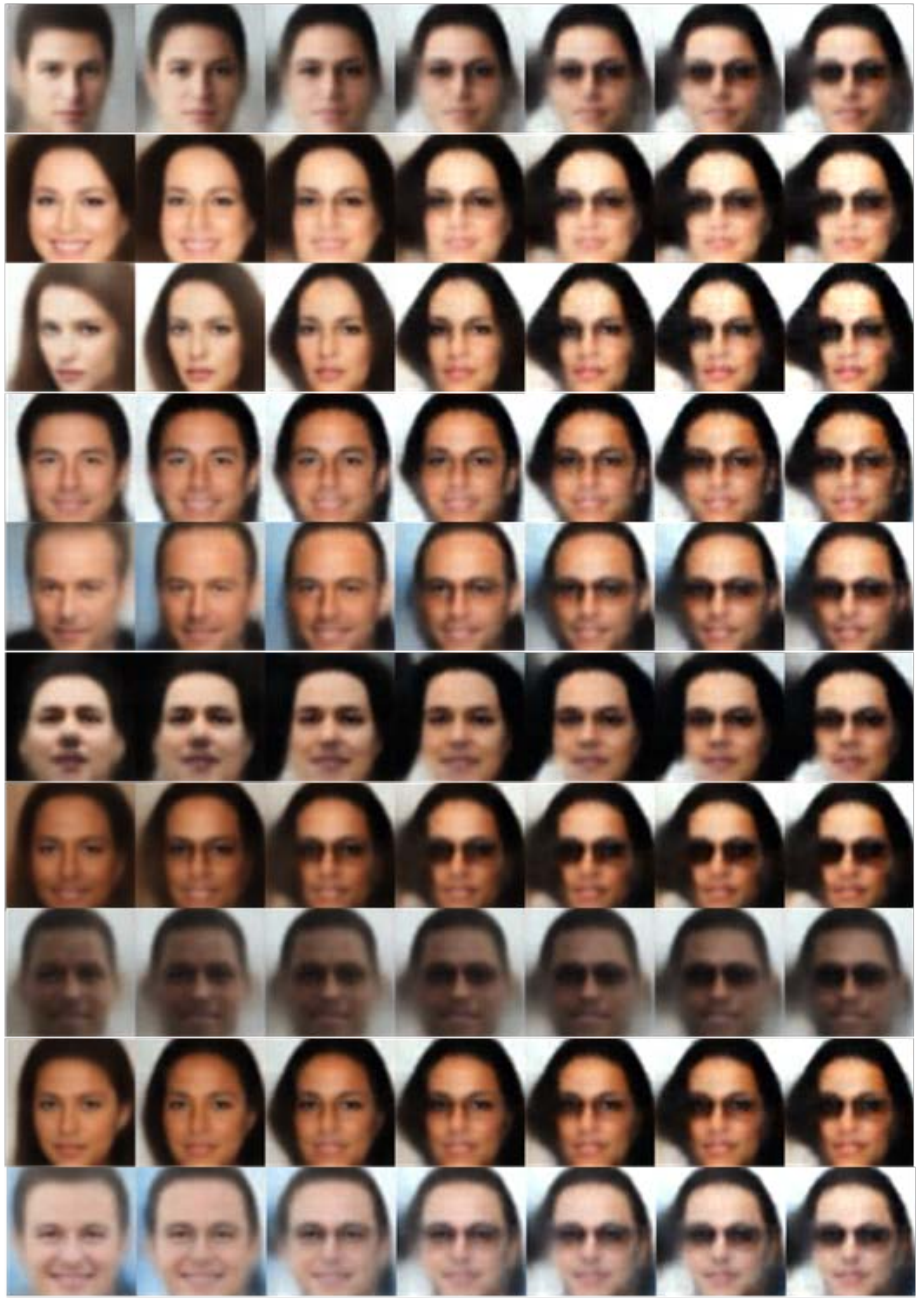}
    \end{minipage}
}
\subfigure[azimuth]
{
 	\begin{minipage}[b]{.3\linewidth}
        \centering
        \includegraphics[scale=0.3]{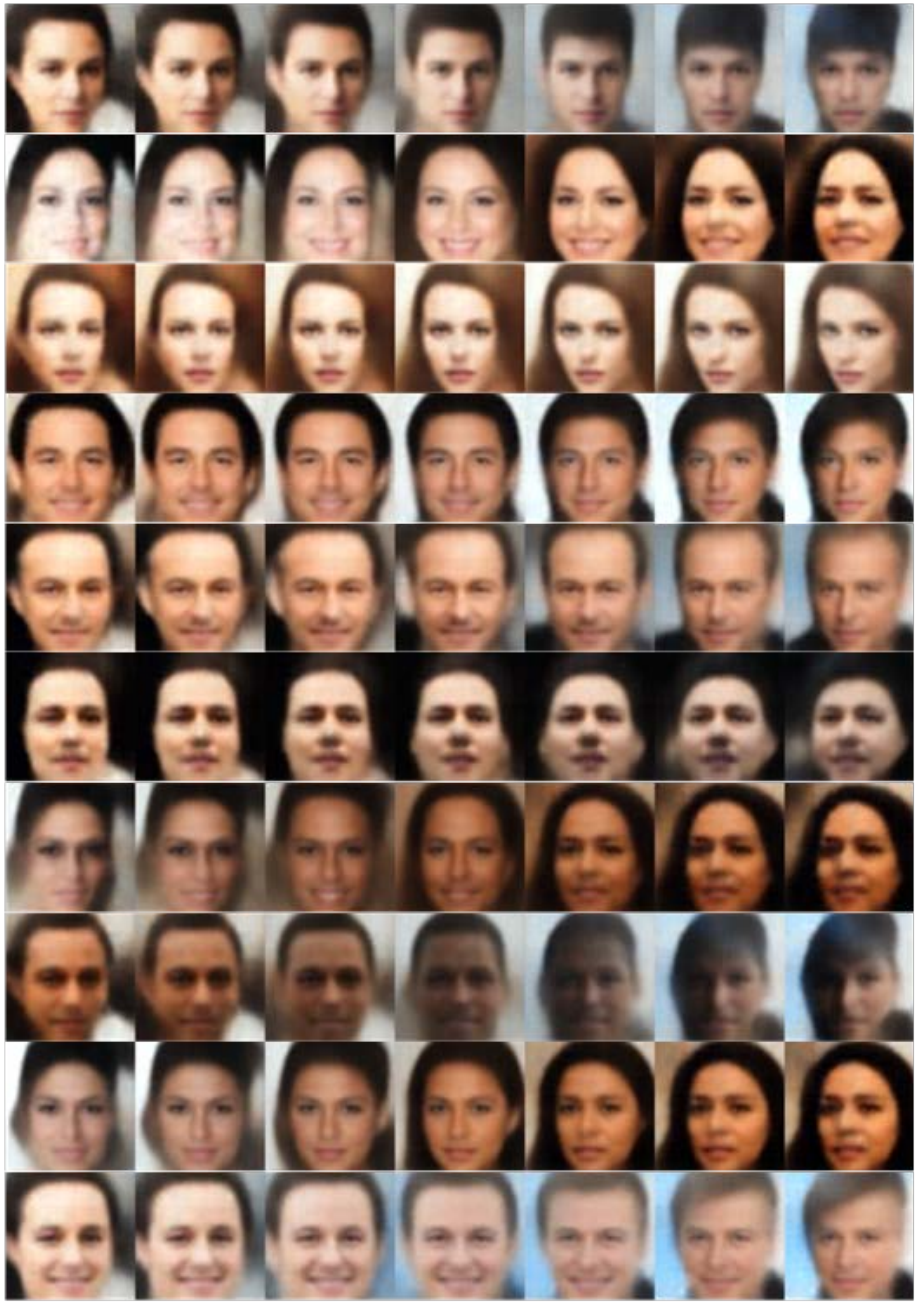}
    \end{minipage}
}
\subfigure[smile]
{
 	\begin{minipage}[b]{.3\linewidth}
        \centering
        \includegraphics[scale=0.3]{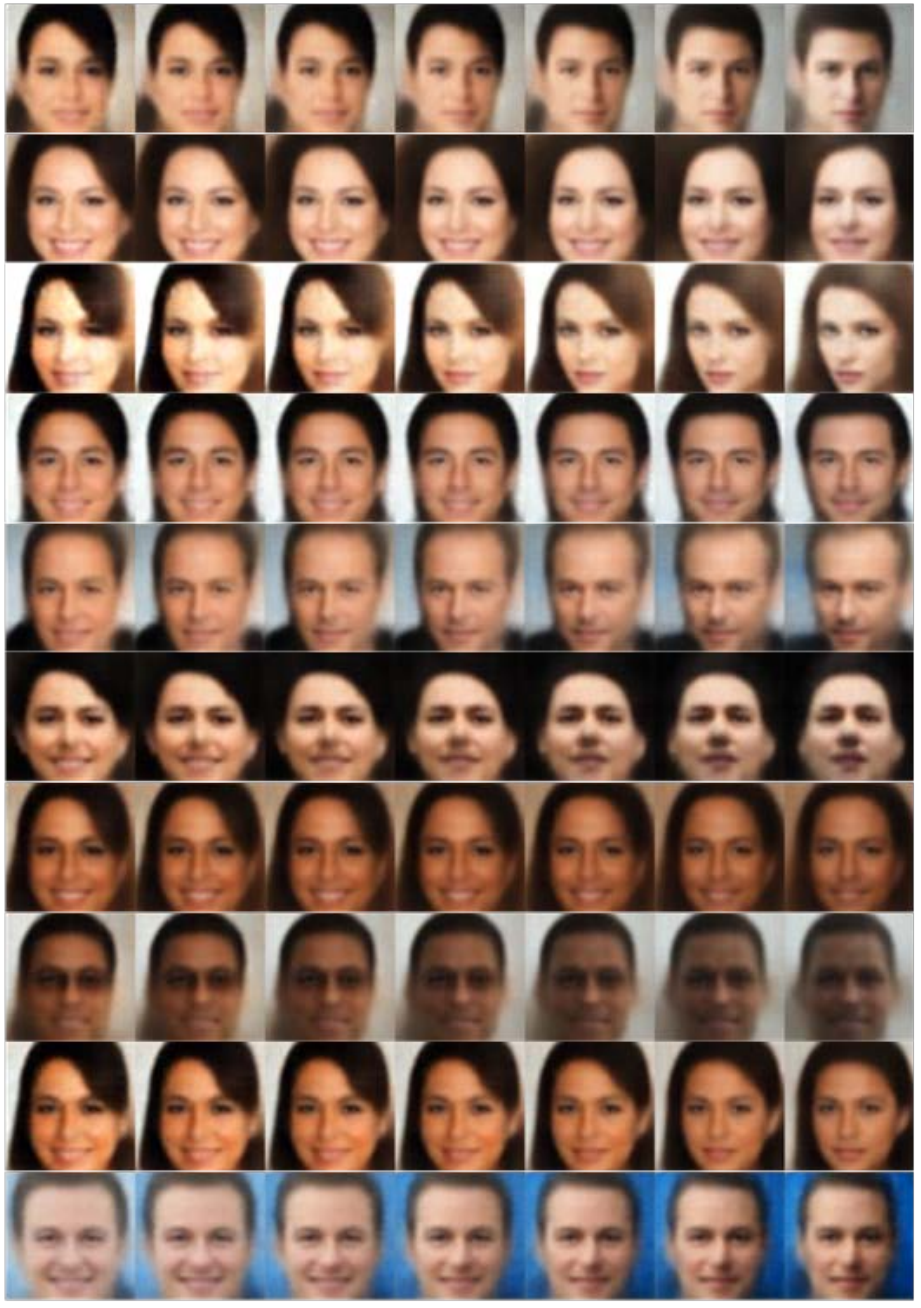}
    \end{minipage}
}
\subfigure[hue]
{
 	\begin{minipage}[b]{.3\linewidth}
        \centering
        \includegraphics[scale=0.3]{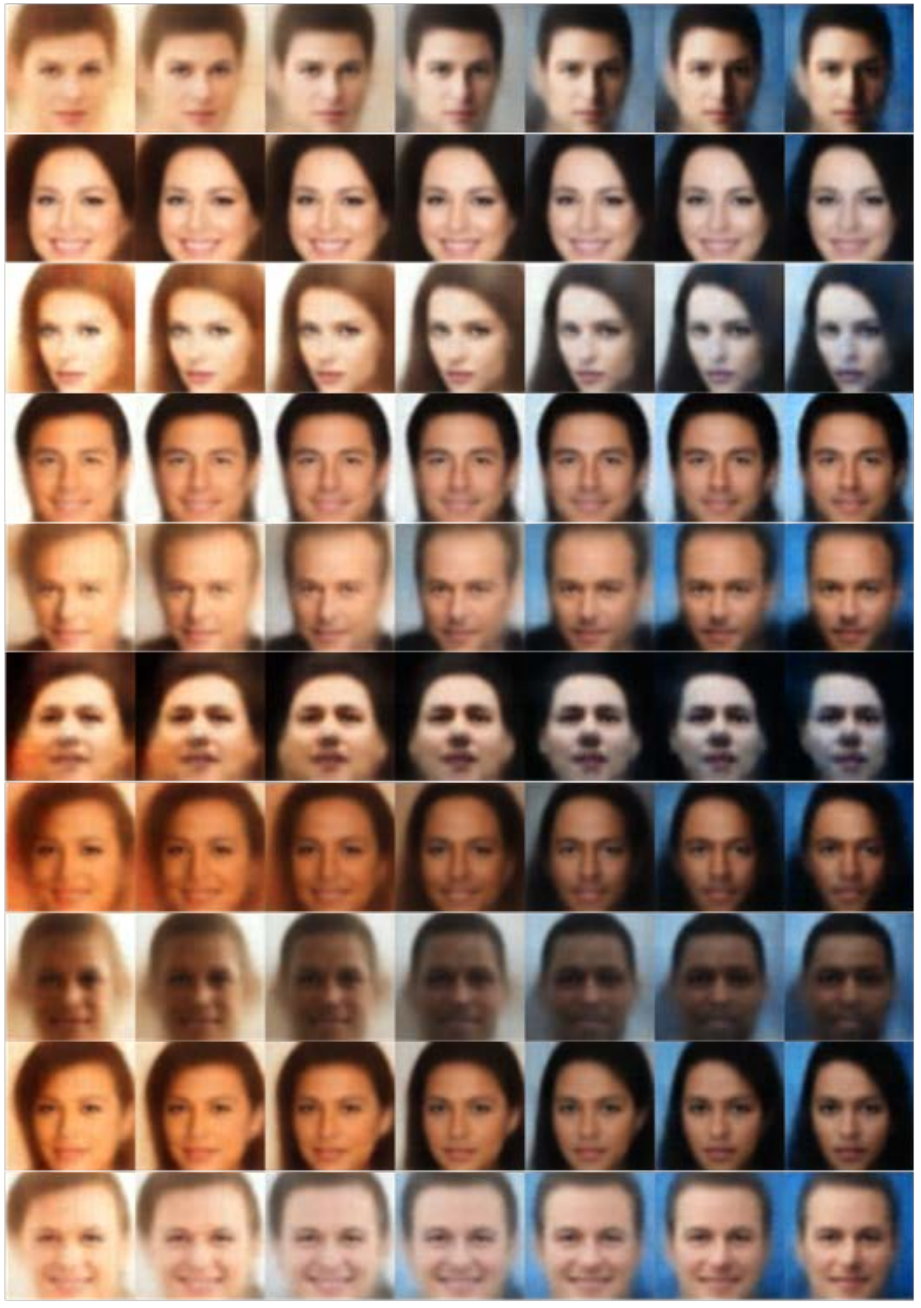}
    \end{minipage}
}
\caption{Disentanglement results for unsupervised $\beta$-CapsNet on CelebA}
\end{figure}

\begin{figure}[!htbp]
	\centering
	\includegraphics[scale=0.4]{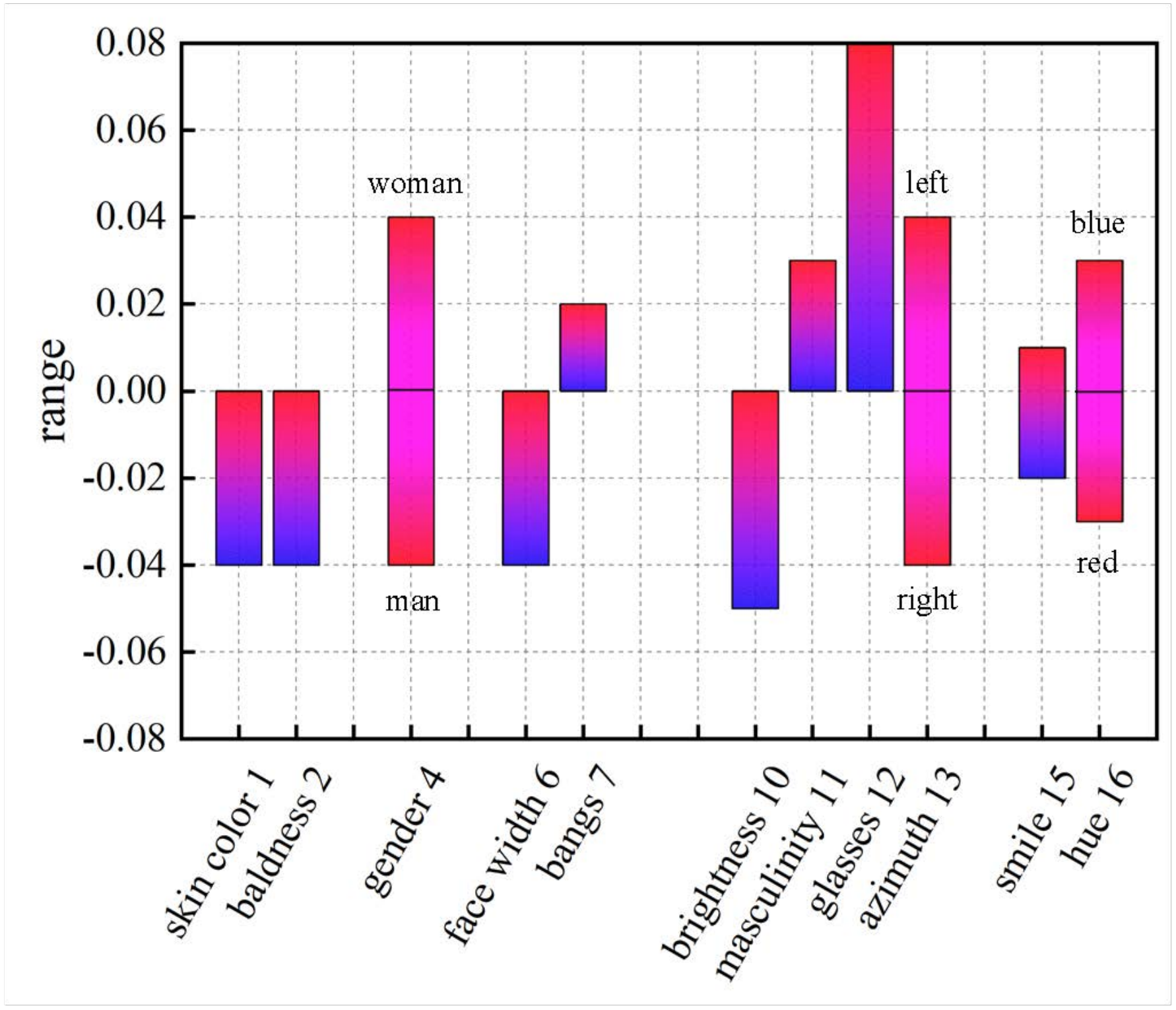}
	\caption{Dimension and its range of disentangled factors on CelebA. The ranges for the components of most disentangled factors are one-sided except for three obviously symmetrical factors such as gender, azimuth and hue}
	\label{fig17}
\end{figure}

\section{Conclusions and Future Work}

In this paper, we have introduced $\beta$-CapsNet, a novel method for learning disentangled representations of CapsNet through variational information bottleneck. Our $\beta$-CapsNet achieves better disentanglement than CapsNet on the MNIST and Fashion-MNIST datasets in a supervised manner. It also learns more interpretable properties than $\beta$-VAE without supervision on unsupervised MNIST \& Fashion-MNIST, 3D chairs and CelebA datasets. We present class independent mask vector, a refinement of the existing mask matrix that helps to learn disentanglement between different categories within a constrained space on supervised datasets. We also propose unsupervised $\beta$-CapsNet and the corresponding dynamic routing algorithm for unsupervised learning.

Variational bound of information bottleneck term in $\beta$-CapsNet is KL divergence between a standard Gaussian and the distribution of capsule representation. We approximate the KL divergence by limiting the mean of capsule to approach 0 due to the uncertainty of the variance, which is a very simple regularization technique and is suitable for almost all networks. Therefore, it is an interesting topic to investigate whether it can help other models to learn disentangled representations. In addition, the reason why restricting the mean and reducing the space of the representations is beneficial to disentangled representations may be a significant direction. Figuring out the problem might give us some insight into the nature of disentanglement.

\bibliography{mybibfile}%参考文献

\end{document}